\documentclass[journal]{IEEEtran}
\usepackage{adjustbox}
\usepackage{array,multirow}
\usepackage[table,xcdraw]{xcolor}
\usepackage{makecell}
\usepackage{float}
\usepackage{comment}
\usepackage{graphicx}
\usepackage{color}
\usepackage{pgfplotstable}
\usepackage{verbatim}
\usepackage{gensymb}
\usepackage{amssymb}
\usepackage{rotating}
\usepackage[normalem]{ulem}
\usepackage{gensymb}
\usepackage{pdflscape}
\usepackage{subcaption}   
\usepackage{amsmath}
\usepackage{algorithmic}
\usepackage{algorithm}
\usepackage{amssymb}
\usepackage{lineno}
\usepackage{flushend}
\usepackage{adjustbox}
\usepackage{ctable} 
\usepackage[font=small,skip=2pt]{caption}
\usepackage{scalerel}
\usepackage{tikz}
\usepackage{soul}
\usepackage{siunitx}

\usepackage{graphicx}
\usepackage{adjustbox}

\newcommand\VRule[1][\arrayrulewidth]{\vrule width #1}
\usepackage[utf8]{inputenc}
\usepackage{mathtools, nccmath}

\usetikzlibrary{svg.path}

\definecolor{orcidlogocol}{HTML}{A6CE39}
\tikzset{
  orcidlogo/.pic={
    \fill[orcidlogocol] svg{M256,128c0,70.7-57.3,128-128,128C57.3,256,0,198.7,0,128C0,57.3,57.3,0,128,0C198.7,0,256,57.3,256,128z};
    \fill[white] svg{M86.3,186.2H70.9V79.1h15.4v48.4V186.2z}
                 svg{M108.9,79.1h41.6c39.6,0,57,28.3,57,53.6c0,27.5-21.5,53.6-56.8,53.6h-41.8V79.1z M124.3,172.4h24.5c34.9,0,42.9-26.5,42.9-39.7c0-21.5-13.7-39.7-43.7-39.7h-23.7V172.4z}
                 svg{M88.7,56.8c0,5.5-4.5,10.1-10.1,10.1c-5.6,0-10.1-4.6-10.1-10.1c0-5.6,4.5-10.1,10.1-10.1C84.2,46.7,88.7,51.3,88.7,56.8z};
  }
}

\newcommand\orcidicon[1]{\href{https://orcid.org/#1}{\mbox{\scalerel*{
\begin{tikzpicture}[yscale=-1,transform shape]
\pic{orcidlogo};
\end{tikzpicture}
}{|}}}}
\usepackage{caption}

\usepackage{hyperref} 

\ifCLASSINFOpdf

\else
 
\fi

\DeclareUnicodeCharacter{2212}{-}
\begin{document}
\bstctlcite{IEEEexample:BSTcontrol}

\title{
Neuromorphic Camera Denoising using Graph Neural Network-driven Transformers
}

\author{

	Yusra~Alkendi$^{1,2}$ \orcidicon{0000-0001-6618-5317},
	Rana~Azzam$^{1}$ \orcidicon{0000-0003-0378-1909},
	Abdulla~Ayyad$^{1}$ \orcidicon{0000-0002-3006-2320},
	Sajid Javed$^{1,3}$ \orcidicon{0000-0002-0036-2875},
    Lakmal~Seneviratne$^{1}$ \orcidicon{0000-0001-6405-8402}, 
	and~Yahya~Zweiri$^{1,2}$ \orcidicon{0000-0003-4331-7254}\

	\thanks{$^{1}$Yusra Alkendi, Rana~Azzam, Abdulla~Ayyad, Sajid~Javed, Lakmal Seneviratne, and Yahya Zweiri are with the Khalifa University Center for Autonomous Robotic Systems (KUCARS), Khalifa University of Science and Technology, Abu Dhabi, UAE, e-mail: {\small \{yusra.alkendi, yahya.zweiri@ku.ac.ae\}}}	

	\thanks{$^{2}$Yusra Alkendi and Yahya Zweiri are also with the Department of Aerospace Engineering, Khalifa University of Science and Technology, Abu Dhabi, UAE.}
	
	\thanks{$^{3}$Sajid Javed is also affiliated with the Department of Electrical Engineering and Computer Science, Khalifa University of Science and Technology, Abu Dhabi, UAE.}
  }
\markboth{IEEE TRANSACTIONS ON NEURAL NETWORKS AND LEARNING SYSTEMS}{Alkendi \MakeLowercase{\textit{et al.}}: Bare Demo of IEEEtran.cls for IEEE Journals}
\maketitle

\begin{abstract}
Neuromorphic vision is a bio-inspired technology that has triggered a paradigm shift in the computer-vision community and is serving as a key-enabler for a wide range of applications.
This technology has offered significant advantages including reduced power consumption, reduced processing needs, and communication speed-ups. 
However, neuromorphic cameras suffer from significant amounts of measurement noise.
This noise deteriorates the performance of neuromorphic event-based perception and navigation algorithms. 
In this paper, we propose a novel noise filtration algorithm to eliminate events which do not represent real log-intensity variations in the observed scene.
We employ a Graph Neural Network (GNN)-driven transformer algorithm, called GNN-Transformer, to classify every active event pixel in the raw stream into real-log intensity variation or noise.
Within the GNN, a message-passing framework, referred to as EventConv, is carried out to reflect the spatiotemporal correlation among the events, while preserving their asynchronous nature.
We also introduce the Known-object Ground-Truth Labeling (KoGTL) approach for generating approximate ground truth labels of event streams under various illumination conditions. 
KoGTL is used to generate labeled datasets, from experiments recorded in challenging lighting conditions, including moon light. 
These datasets are used to train and extensively test our proposed algorithm.
\textcolor{black}{When tested on unseen datasets, the proposed algorithm outperforms state-of-the-art methods by at least 8.8\% in terms of filtration accuracy.}
Additional tests are also conducted on publicly available datasets (ETH-Zurich \textit{Color-DAVIS346} datasets) to demonstrate the generalization capabilities of the proposed algorithm in the presence of illumination variations and different motion dynamics.
Compared to state-of-the-art solutions, qualitative results verified the superior capability of the proposed algorithm to eliminate noise while preserving meaningful events in the scene.

\end{abstract}


\IEEEpeerreviewmaketitle

\section{Introduction}
\IEEEPARstart{O}\small{v}er the last decade, advances in image sensor technologies have rapidly progressed, providing several alternative solutions for scene perception and navigation.
The neuromorphic event-based camera also known as Dynamic Vision Sensor (DVS) is an asynchronous sensor that mimics the neurobiological architecture of the human retina. 
It has caused a paradigm shift in vision algorithms due to the way visual data is acquired and processed.
Instead of capturing image frames as conventional cameras, event-based cameras report asynchronous temporal differences in the scene and form a continuous stream of events which is generated when the log-intensity of each pixel changes (i.e. events) in the order of microseconds ($\mu$s).
The event-based camera has the capability to overcome the limitations of conventional cameras by providing data at low latency (20 $\mu$s), high temporal resolution ($>$800kHz), high dynamic range (120 dB), and no motion blur \cite{Event_based_Vision_A_Survey}.
These sensors are able to operate in a wide range of challenging illumination environments (i.e. low light conditions), while consuming an extremely low amount of power e.g., 10−30 mW \cite{Event_based_Vision_A_Survey}.

Recently, event-based cameras have been successfully employed to perform challenging tasks such as object tracking \cite{s13_Event_driven_ball}, object recognition \cite{s17_Fully_Event_Based}, monitoring \cite{s14_Bauer2007}, depth estimation \cite{s19_Rebecq2018}, optical flow estimation \cite{s22-Zhu-RSS-18}, high dynamic range (HDR) image reconstruction \cite{s8_rebecq2019high}, segmentation \cite{s129_segmentation}, guidance \cite{ours1,ours2}, and simultaneous localization and mapping (SLAM) \cite{s27_UltimateSlam}.
In the literature, the performance of such event-based applications degrades in the presence of noise \cite{Event_based_Vision_A_Survey}.
The noise associated with the generated event data using  DVS could be due to the lighting conditions, motion dynamics in the scene, or the sensor parameters. 
Extraction of meaningful event data in presence of noise is considered a major challenge and needs further developments as mentioned in \cite{Event_based_Vision_A_Survey}.

In poor lighting conditions, events corresponding to features or edges of moving objects are highly scattered and an overwhelming amount of noise is present even if optimal camera parameters are used \cite{s27_UltimateSlam,ours1}.
Due to the humongous amounts of events generated by DVS, manually identifying and filtering noise out is a challenging task and therefore research efforts are needed especially  towards noise identification and filtration in the presence of challenging lighting variations. 
To date, a mathematical model that accurately describes the noise associated with event streams is not yet formulated. 
To circumvent such challenge, machine learning approaches can be employed to approximately model and characterize the noise parameters and consequently filter out events that do not correspond to real intensity variations in the scene. 
However, the lack of labeled datasets to train event-denoising models has hindered the progress of machine learning solutions to this problem. 
In this paper, we propose Known-object Ground-Truth labeling (KoGTL) approach for generating approximate ground truth labels for event streams. This is directed towards developing an ML-based event denoising technique that inherently copes with the nonlinear behavior of the noise associated with events. 

\begin{figure}
\centering
\centering
\begin{adjustbox}{width=0.5\textwidth}
  \centering
\begin{tabular}{cc}


 

\includegraphics[width=\columnwidth]{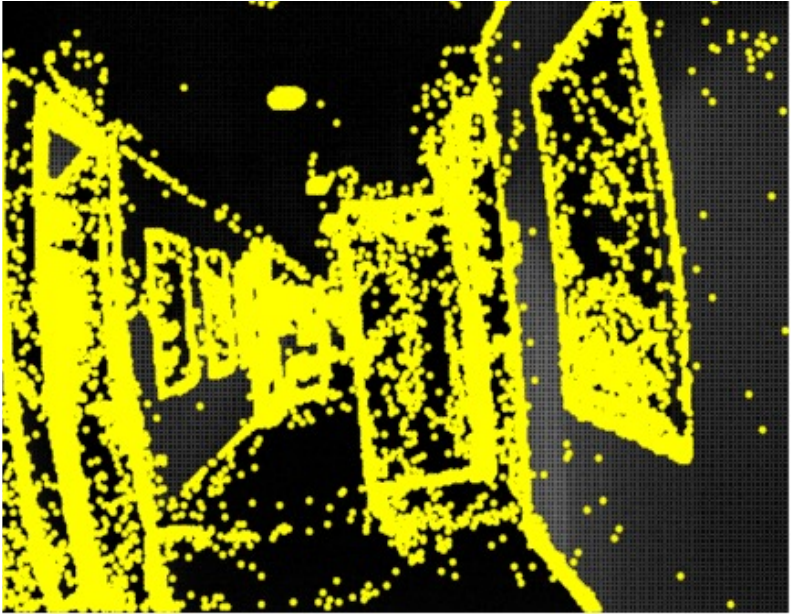} 

&  \includegraphics[width=\columnwidth	]{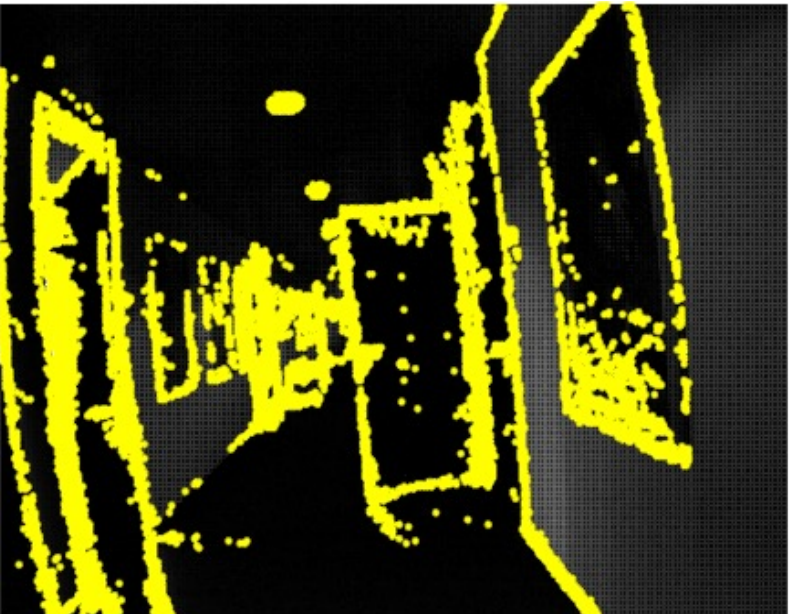}    \\   



\\  

          \centering\LARGE	 (a) Raw Data - DVS Sensor &  \centering\Huge  
        \centering\LARGE	{(b) GNN-Transformer (ours)} 

\end{tabular}
\end{adjustbox}
\caption{Denoising results using \textit{IndoorsCorridor} publicly available dataset in low light scenario \cite{publicDataset}. 
Events (yellow dots) are overlaid on the corresponding APS image for visualization. (a) Raw DVS stream of events and (b) Denoised events using the proposed learning-based method (GNN-Transformer). 
Our GNN-Transformer performs a binary classification to distinguish between actual DVS events and noise. Note that our proposed algorithm does not use APS images for denoising. \textcolor{black}{All events that do not correspond to edges but are visible in the APS image have been filtered out}. Our GNN-Transformer performs significantly better than the state-of-the art methods in challenging lighting conditions (i.e. low light).}\label{qual_5}
\end{figure}

Graph neural networks (GNNs) have shown excellent progress in a plethora of applications \cite{ji2021survey, liu2021toward}.
GNN operates on data structures in the non-Euclidean domain and hence it is considered as part of the geometric deep learning framework.
Particularly, GNNs operate on graphs that model a group of objects referred to as nodes and their relationships, which are referred to as edges \cite{9046288}.
Such data structures are not supported by conventional deep neural networks (DNNs), convolutional neural networks (CNNs), or recurrent neural networks (RNNs).
GNN preserves the structure of the input graph and exploits the knowledge of the dependencies between the nodes to infer knowledge about the data. Hence, we exploit this feature of GNN and propose to design a message passing GNN model that can operate on event streams, preserve the asynchronous nature of events, and learn to solely outflow the noise-free DVS events.

Recently, the transformers have attained significant attention in the machine learning community \cite{khan2021transformers}.
Vaswani \textit{et al.} proposed to model sequence-to-sequence learning task using transformer \cite{vaswani2017attention}.
The self-attention mechanism within the transformer captures the relationships between input and output data and supports parallel processing of sequence recurrent networks. 
Transformers have recently been employed in many applications including natural language processing and computer vision to name a few \cite{kalyan2021ammus,khan2021transformers, chang2021end}.
In this work, we employ transformers within the proposed GNN for the task of identifying and eliminating the noise associated with events generated by DVS. 
To the best of our knowledge, no such research study exists in the literature where GNNs are employed together with transformers for event-based applications.

We propose a novel event denoising (ED) model that can learn spatiotemporal correlations between newly arrived events and the previous active events in the same neighborhood.
This is achieved by means of a GNN-Transformer that operates on event streams encoded into graph structures. 
Our proposed algorithm consists of a message-passing GNN model and a transformer network to perform binary classification of events into real activity events or noise. 
The proposed GNN-Transformer based ED algorithm has the following advantages: (I) It can seamlessly operate on raw event streams without any data preprocessing or camera parameters' tuning , (II) It can efficiently perform in illumination conditions ranging from good light conditions to near darkness conditions, and (III) It shows robustness against different motion dynamics.
\textcolor{black}{The proposed GNN-Transformer is an accurate and general learning-based spatiotemporal event filter that outperforms existing denoising methods \cite{D9_Baldwin2020a,D6_Feng2020,D7_Khodamoradi2017,D1_Liu2015,D2_Padala2018} in various testing scenarios.} Through several tests on publicly available datasets \cite{publicDataset}, the proposed model has proven its effectiveness and capability to denoise incoming streams of events under challenging conditions in terms of illuminations and motion dynamics.
Fig. \ref{qual_5} shows sample denoising results obtained when our proposed algorithm was used on a publicly available dataset recorded in low light conditions \cite{publicDataset}.   
Our proposed algorithm operates on event graphs constructed from the incoming raw event streams where nodes represent the event properties (pixel location and time of arrival). 
The node of interest, i.e. the event that has just been observed, is connected through edges to the rest of the nodes that represent recent activity in the neighborhood. Then, node features are processed to generate seven messages that are sent out along the graph edges in preparation for inference and event classification.
Messages are then aggregated to form a graph signature, based on which the node of interest is classified into real-activity event or noise. 
Since classification is done based on the graph signature rather than the raw node features, the proposed algorithm has achieved generalization across various testing datasets.

To train and test the proposed model, we develop an experimental protocol to acquire event streams from motion in different directions and under various lighting conditions. 
The proposed KoGTL approach is used to label events as real activity events (class 1) or noise (class 0).
The training dataset is then constructed using graph samples that encode event features and neighborhood properties, and their corresponding labels generated using KoGTL. 
It is worth noting that the proposed algorithm accepts input graphs of variable sizes, i.e. varying number of events in a particular spatiotemporal neighborhood.
This property of the proposed ED method is very crucial since it allows for coping with the asynchronous nature of event acquisition.
Experimental evaluations on various training and testing datasets demonstrate excellent performance of the proposed algorithm compared to the existing state-of-the-art methods.
The main contributions of the this work are as follows:

\begin{enumerate}
\item We introduce a novel Known-object Ground-Truth Labeling (KoGTL) approach 
to generate a labeled dataset of noise and real-activity events. This dataset includes varied lighting conditions and relative motions in the visual scene.
\item We design a novel message passing framework, dubbed EventConv, on graphs constructed from DVS events. 
Messages encapsulate the spatiotemporal properties of events in a neighborhood while accounting for the asynchronous nature of data acquisition.
\item  We develop a novel Event Denoising GNN-Transformer architecture based on the novel EventConv layer to distinguish between real-activity and noise events. 
\item We perform extensive evaluations of the proposed algorithm on our labeled dataset and other publicly available event datasets. 
Experiments are conducted to validate the proposed model's generalization capabilities on unseen data involving different motion dynamics and challenging lighting conditions.
\item We release a new dataset (\textit{ED-KoGTL}) with labelled neuromorphic camera events acquired from motions in different directions and under various illumination conditions.
Our labeled dataset is publicly available to the research community $<$\url{https://github.com/Yusra-alkendi/ED-KoGTL}$>$ for benchmark comparison.

\end{enumerate}

The rest of this paper is organized as follows. 
In Section \ref{sec:related_work}, we review related work.
In Section \ref{sec:Proposed Framework}, we describe the proposed algorithm and dataset in detail.
The experimental results are presented in Section \ref{sec:results}.
Finally, the conclusions are drawn in Section \ref{sec:conclusion}.

\section{Background and Related Works}
\label{sec:related_work}

\subsection{Event denoising}
\textcolor{black}{The importance of the event denoising module to event-based computer vision algorithms has been demonstrated through several research work, such as for object recognition \cite{L1_eventGait}, object tracking \cite{L3}, image reconstruction \cite{L3}, and segmentation \cite{L4_AAAI}}. 
DVS produces noise due to various reasons. 
Noise could be generated due to thermal noise and junction leakage currents under constant lighting conditions. This type of noise is referred to as background activity noise. False negative events also generate noise and occur when there is no change in the log intensity. 
Furthermore, 
when a sudden change in illumination happens, a huge amount of random noise occurs in the event stream\textcolor{black}{.}

\begin{figure}
\centering
 \includegraphics[width=0.5\textwidth]{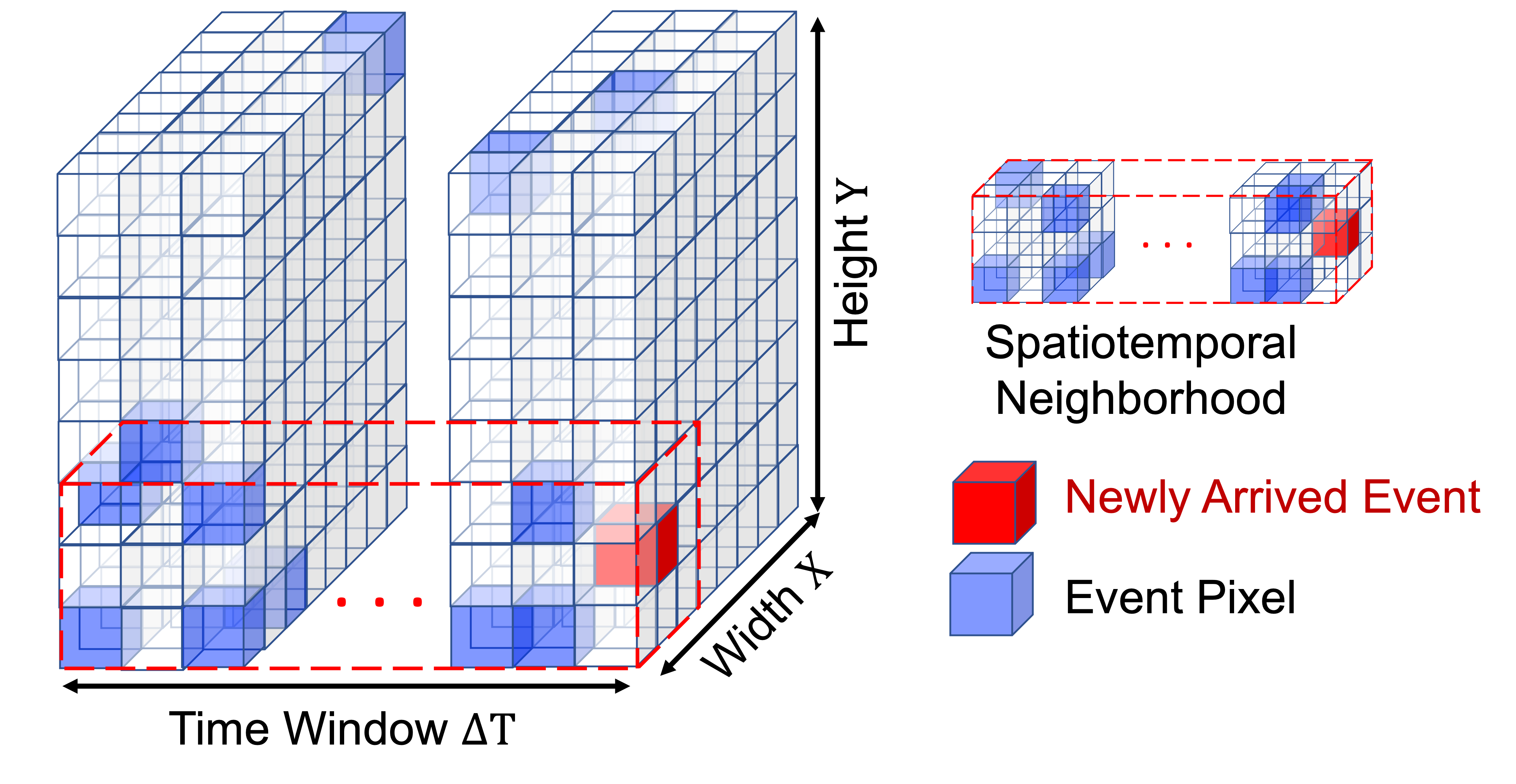}
  \caption{An example of event spatiotemporal neighborhood.}
  \label{fig:dvs_presentation}
\end{figure}

The background activity (BA) events differ from real activity events.
BA lacks temporal correlation with the newly arrived events in the spatial neighborhood while real activity events show meaningful correlation.
Several event noise reduction methods have been proposed in the literature. \textcolor{black}{These methods can be categorized into conventional methods \cite{D0_Delbruck2008,D1_Liu2015,D7_Khodamoradi2017,D6_Feng2020,D12_Guo2020,D12_Guo2020b} and deep learning methods \cite{D9_Baldwin2020a,L5_TORE_baldwin, L3}.} The most widely prevalent filtering approach is based on the nearest neighbor (NNb) method and hence on spatiotemporal correlation \cite{D0_Delbruck2008,D1_Liu2015,D7_Khodamoradi2017}.
In such filters, the properties of the previously generated events in a spatiotemporal neighborhood are utilized to determine if a newly arrived event represents real activity. The parameters of the spatiotemporal window have to be tuned by the user.
Fig. \ref{fig:dvs_presentation} shows the representation of event spatiotemporal neighborhood, where the newly arrived event data at $t_i$ is marked as a red pixel and its spatial neighborhood is shown in blue. 
Therefore, such approaches require additional memory resources to retain the previous and the newly arrived events' properties for processing.
\setlength{\textfloatsep}{2pt}

The BA filter proposed by Delbruck \cite{D0_Delbruck2008} classifies events that have less than eight other events in their spatiotemporal neighborhood as noise. 
One drawback of such approach is observed when two BAs are close enough in one spatiotemporal region where the filter would consider them as real activity events. 
Furthermore, Liu et al. \cite{D1_Liu2015} have proposed a filter to tackle the problem of increased memory requirements by sub-sampling pixels into groups, where instead of projecting every pixel into a memory cell, one memory cell would hold a sub-sampled group of pixels. 
The filtration accuracy relies heavily on the sub-sampling factor, where the filtration accuracy decreases when the sub-sampling factor is greater than 2.

\begin{figure}
\centering
 \includegraphics[width=0.5\textwidth]{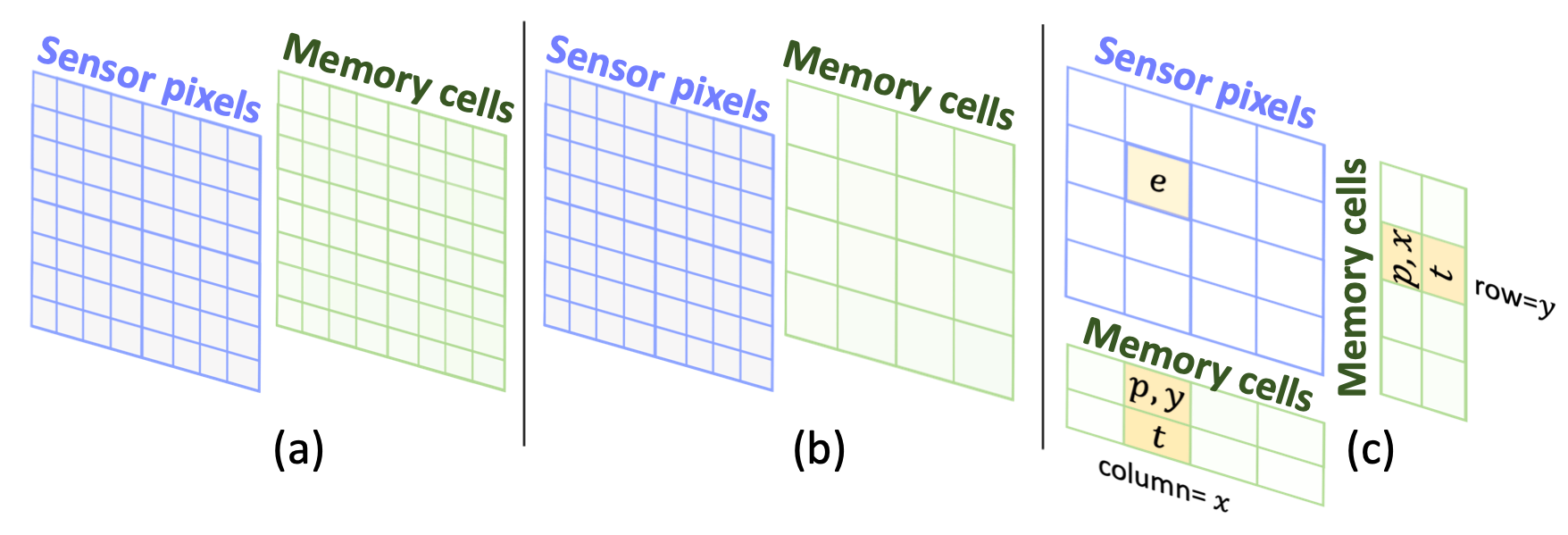}
  \caption{Examples of memory strategy of different  spatiotemporal filters \cite{D7_Khodamoradi2017}: a) shows one memory cell per pixel \cite{D0_Delbruck2008}, b) shows one memory cell per two sub-sampling group \cite {D1_Liu2015}, and c) shows two memory cells for each column and row \cite{D7_Khodamoradi2017}.}
  \label{fig:BAmethods}
\end{figure}

Khodamoradi and Kastner proposed another storage technique for events and their timestamps to utilize less memory space \cite{D7_Khodamoradi2017}. 
Particularly, the most recent event in every row and column is stored along with its corresponding polarity and timestamp into two 32-bit memory cells. Hence, if two events are acquired in the same column, but two different rows, within a short temporal window, the recent event will override the old one in the memory. This is a serious limitation of this approach as establishing spatial correlation is deemed impossible, and thereby more real activity events could be sorted out as noise. 
Fig. \ref{fig:BAmethods} depicts the techniques used to store events in the memory prior to filteration as proposed in \cite{D0_Delbruck2008, D7_Khodamoradi2017, D1_Liu2015}. 
\setlength{\textfloatsep}{2pt}

To overcome memory and computational complexity issues, Yang \textit{et al.} proposed a density matrix in which each arriving event is projected into its spatiotemporal region \cite{D6_Feng2020}. The denoising process in this method consists of two steps; (1) removing random noises and (2) removing hot pixels (permanent active or permanent broken event pixels). \textcolor{black}{Moving to the learning-based denoising approaches, in the literature, Baldwin \textit{et al.} \cite{D9_Baldwin2020a, L5_TORE_baldwin} and Duan \textit{et al.} \cite{L3} have proposed a convolutional neural network and U-net network to filter DVS noises, respectively. } 

\textcolor{black}{It is also evident that the performance of the existing denoising methods rely on tunable parameters e.g., spatiotemporal window size, event camera settings, environmental illumination conditions, and camera motion dynamics \cite{D6_Feng2020, D7_Khodamoradi2017, D9_Baldwin2020a,L5_TORE_baldwin, L3}. Such parameters are application-dependent and manually tuning them may lead to satisfactory denoising results, especially in good lighting conditions. }
Despite setting the camera parameters to their optimal values though, features or edges of moving objects in very low illumination conditions are highly scattered and very noisy. 
In order to extract meaningful information from varying light conditions, the need for a method that can reject these noises and sharpen the real event data is essential. 
Nevertheless, spatiotemporal correlation-based and deep learning methods of event denoising remain largely unexplored.

 \vspace{-2mm}

\subsection{Graph Neural Networks and Transformers}
Graph Neural Networks (GNNs) are deep learning models that operate on non-Euclidean data structures such as graphs. 
GNNs take into account the properties of each graph node and its connectivity within its neighborhood, regardless of the order in which data is provided to the neural network.
It is also worth mentioning that the size of the input graph could be variable for the same network which makes GNN very well-suited for the application in hand. 
Owing to its expressive power and model flexibility, GNN has recently been employed in a wide range of applications e.g., visual understanding on images \cite{ GNN3, GTNNLS}).
Interested readers can explore more details in this direction in these recent surveys \cite{zhou2020graph, chen2020bridging}.

\begin{figure*}[t]
\centering
 \includegraphics[width=0.92\textwidth]{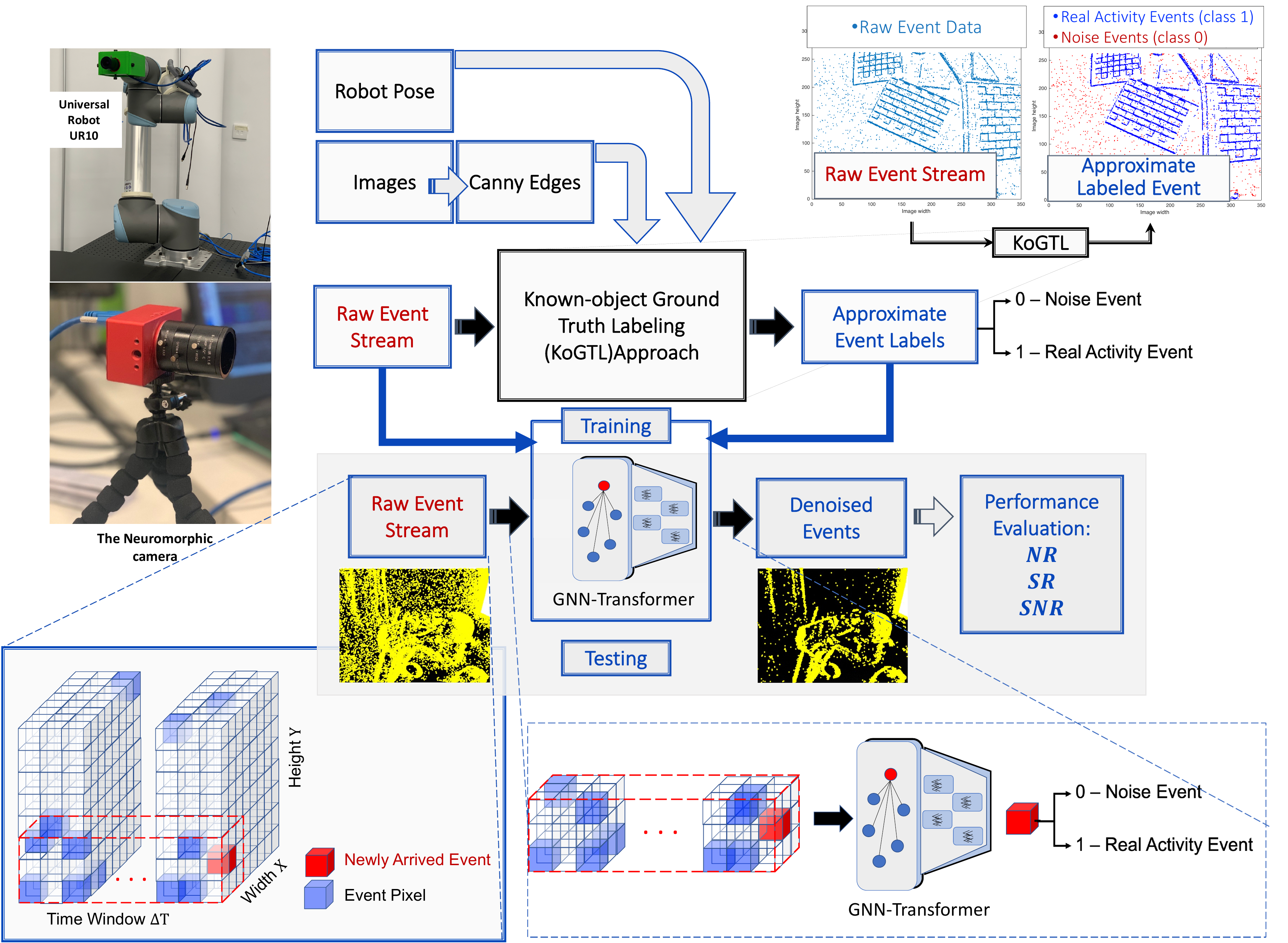}
  \caption{Proposed event denoising framework. A GNN-Transformer based event denoising algorithm is developed and trained on event datasets, generated and labeled using the proposed Known-object Ground-Truth Labeling (KoGTL) approach. The proposed algorithm classifies incoming event streams into real activity events or noise. }
  \label{fig:dvs_proposedApproach1}
\end{figure*}

\setlength{\textfloatsep}{2pt}

There are different types of graph representations exhibiting various levels of complexity (i.e. number of connections and dimension) 
to address the problem in question. 
For instance, the work proposed in \cite{pointcloudGNN} and \cite{Rana21} designed graphs to represent point-clouds and ground vehicle poses, respectively. The features of the nodes and edges in each graph encode information necessary to perform the problem in hand, like the point 3D coordinates and the 2D pose of the robot. 
In \cite{pointcloudGNN}, a stack of EdgeConv layers is proposed to capture and exploit fine-grained geometric properties of point clouds which are then employed to carry out classification and segmentation for point cloud data.
Another graph convolutional layer is proposed in \cite{Rana21}, called PoseConv, to carry out global optimality verification of 2D pose graph SLAM. 

There are several types of GNNs, designed to fit different graph structures for different tasks. 
Our proposed algorithm adopts a message passing algorithm on graphs, which is carried out in two stages: message passing and aggregation \cite{zhou2020graph}. 
To construct a graph with a unique signature that reflects the nature of input data, in this work, spatiotemporal correlation functions are used. This is to reflect the nonlinear nature of the noise associated with DVS event streams.
In addition, the graph isomorphism problem might occur when two different graphs might have an identical representation when reduced by the aggregation function. Inspired by \cite{mansour2017message}, we employ a nonlinear activation within the aggregation stage to handle the graph isomorphism issue.
This is to generate a unique graph signature to represent the spatiotemporal correlation between the nodes of the constructed graphs.

Recently, transformers have demonstrated state-of-the-art performance on a multitude of applications including natural language processing \cite{kalyan2021ammus} and vision systems \cite{khan2021transformers,carion2020end,lee2019tetris}. 
The self-attention head captures the relationship between inputs and outputs and supports parallel processing of sequential recurrent networks. 
In this paper, we demonstrate the scalability of transformers on neuromorphic vision sensors and their capability to handle the asynchronous nature of events. 
This is designed within the graph layer that employs a message passing algorithm to process the dynamic and variant nature of event streams.
The output of the graph is then processed by the transformer, \textcolor{black}{prior} to the final classification stage which removes noise from the event stream.

\section{Proposed Framework}
\label{sec:Proposed Framework}
In this paper, a novel GNN-Transformer is proposed and trained to predict if an incoming DVS event represents noise or a real log-intensity variation in the scene. 
Real log-intensity variation is a representation of a meaningful feature within the scene e.g., the edge of an object. 
The overall framework of the proposed event denoising algorithm is illustrated in Fig. \ref{fig:dvs_proposedApproach1}. 
In the below subsections, we explain each component in detail.
 \vspace{-2.5mm}

\subsection{Known-object Ground-Truth Labeling (KoGTL)} \label{Sec:KoGTL}
The availability of labeled datasets is key to the success of supervised learning algorithms. 
To that end, we propose a novel offline methodology, referred to as Known-object Ground-Truth Labeling (KoGTL) which classifies DVS event stream into two main classes: real or noise event. 
We use KoGTL to generate labeled datasets and train a neural network to predict whether an event represents noise or real activity in the scene. 

\subsubsection{Experimental setup} \label{setup}
The main idea behind the KoGTL is to use a multi-trial experimental approach to record event streams and then perform labeling. 
More specifically, a dynamic active pixel vision sensor (DAVIS346C) is mounted on a Universal Robot UR10 6-DOF arm \cite{ur10e}, in a front forward position and repeatedly moved along a certain (identical) trajectory under various illumination conditions. The UR10 manipulator ensures a repeatability margin of 100 microns along a trajectory, when performed repeatedly.
The DAVIS346C provides a spatial resolution of 346$\times$260, minimum latency of 12 $\mu$s, band-width of 12 MEvent/s and a dynamic range of 120 dB \cite{davis346c}. 
The events are recorded along with two other measurements: (1) the camera pose at which the data was recorded, which we obtain through kinematics of the robot arm and (2) the intensity measurements from the scene obtained using the augmented active pixel sensor which are referred to as APS images hereafter. 

Four experimental scenarios are adopted where data is acquired from repeated transnational motion of the robot along square trajectories under different lighting conditions; particularly $\thicksim$750lux, $\thicksim$350lux, $\thicksim$5lux, and $\thicksim$0.15lux. 
Streams of events with the corresponding APS images and robot poses were acquired for about five seconds per experimental scenario. 
Although the camera motion is identical in all experiments and the depicted scene (APS image) does not change, the properties of the event streams vary due to changes in illumination. 
Two of the experimental scenarios are used for training the proposed event denoising method, while the other two are used exclusively for testing and model evaluation.

\subsubsection{Labeling Framework}\label{labeling}
\begin{figure}
\centering
 \includegraphics[width=0.48\textwidth]{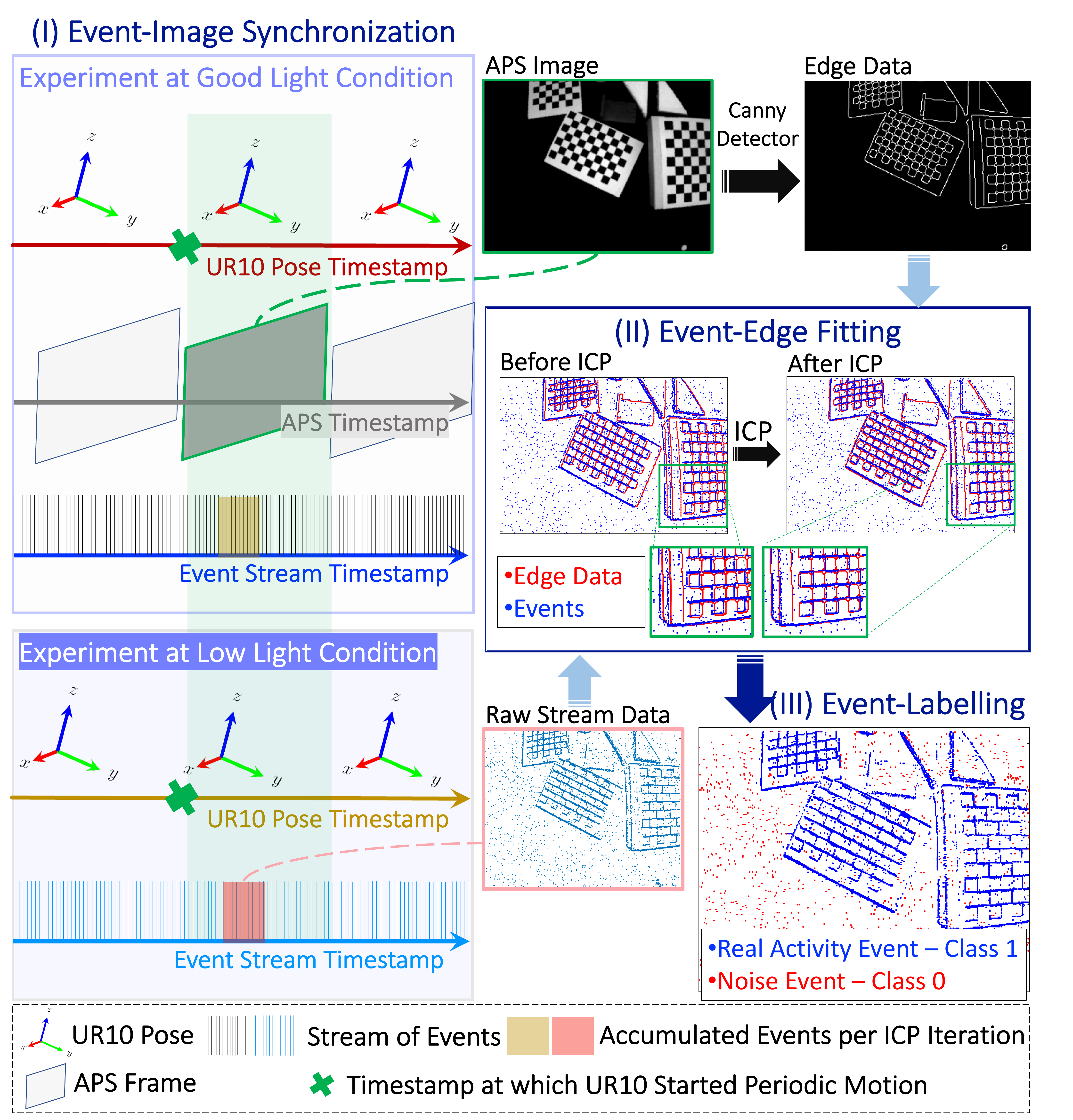}
  \caption{KoGTL labeling framework. KoGTL is a novel DVS event labeling methodology developed to classify DVS events, acquired under various illumination conditions, into two main classes: real event or noise. The proposed KoGTL labels events that are acquired using a multi-trial experimental approach, along with two measurements, camera pose and intensity measurements of the scene.}
  \label{fig:KoGTL}
\end{figure}

The proposed KoGTL labeling algorithm is divided into three main stages including Event-Image Synchronization, Event-Edge Fitting and Event-Labeling as depicted in Fig. \ref{fig:KoGTL}. 

\noindent\textbf{Event-Image Synchronization:} All the recorded experiments are first synchronized based on the time at which the robot arm has started moving (Fig. \ref{fig:KoGTL}-(I)).
Consequently, following identical camera trajectories allows for synchronizing events and APS images across different lighting conditions. 
More specifically, events recorded under poor lighting conditions can be overlaid on APS images captured at the same camera pose under good lighting conditions given that the scene is identical across all experiments. 
This facilitates matching events recorded in low-lighting conditions to alternative APS image features representing the same scene, which is extremely crucial for the success of the second stage. 
This would not have been possible using the APS images captured in low-lighting conditions where variations in intensities and hence features (edges) from the scene are absent.

\noindent\textbf{Event-Edge Fitting:} In the second stage, Canny edge detector \cite{canny1986computational} is used to extract edges from the APS images captured along the trajectory under good lighting conditions. 
The events captured between two consecutive APS images ($t_{APS,i}$ $<=$ $t_{event}$ $<$ $t_{APS,{i+1}}$), are accumulated for every lighting scenario forming a 2D vector as depicted in Fig. \ref{fig:KoGTL}-(I). 
Using the iterative closest point (ICP) fitting technique \cite{icp}, event data are fitted to their corresponding APS edge data. 
Fitting was done in several stages because of the high temporal resolution of DVS data acquisition. 
Events might slightly deviate from APS edges due to imperfections in the time-synchronization of events and APS data. Therefore, ICP is used to perfectly overlay them and correct any resulting spatial shift as shown in Fig. \ref{fig:KoGTL}-(II).

\noindent\textbf{Event-Labeling:} In the third stage, events that were fitted to edges in the APS images are labeled as real-activity events (Class 1), as shown in Fig. \ref{fig:KoGTL}-(III). Other events that fall out of a spatial window around edge pixels (between $+B$ and $-B$ pixels) are considered noise (Class 0). For our dataset, events are classified as noise when they are more than two pixels away (i.e. $B=2$) from an edge in the APS image. This window size was selected based on visual observation of the fitting results using multiple $B$ values.


\subsection{Proposed GNN-Transformer Algorithm for Event Denoising} \label{graphtransfomer}

In this section, we explain the proposed GNN-Transformer for event denoising as depicted in Fig. \ref{fig:dvs_proposedApproach}. 
GNN-Transformer consists of three main stages: event graph construction, message passing on graphs, and event classification. 

\subsubsection{Event Graph Construction} 
Unlike conventional image frames, event data arrives asynchronously within a spatial resolution of $H \times W$ pixels (Fig. \ref{fig:dvs_proposedApproach}-I). Every pixel encodes log intensity variations in the visual scene and is represented by a tuple $e$ $=<$$x$, $y$, $t$, $p$$>$, where ($x$, $y$) are the pixel coordinates at which an event occurred, $t$ is the event's timestamp, and $p$ is the event's polarity (either 1 or -1, signifying an increase or a decrease in the intensity, respectively). 
A sequence of events within a spatiotemporal neighborhood is referred to as a local volume. 
The local volume is defined in terms of its spatial ($L \times L$) and temporal ($T$) dimensions around the event of interest. For example, if $L=1$ and $T=1$, the local volume includes the events arriving in a spatial window of 3$\times$3 pixels around the event of interest in the previous 1 $ms$. 

\textcolor{black}{When a new event arrives, $e_i$ (Fig. \ref{fig:dvs_proposedApproach}-II), a graph $G$ that represents the local volume of the event is constructed (Fig. \ref{fig:dvs_proposedApproach}-III). The nodes of the graph are all the events in the defined local volume. 
Every node has three features $<$$(x_j), (y_j), (t_j)$$>$, where $j$ is a node in the graph, $x_j$, $y_j$ are the pixel coordinates at which the event occurred and $t_j$ is the event's timestamp.} In this work, we omit the use of event polarity as a node feature because of the fact that event polarity is affected by the sensitivity of events to changes in scene illumination which may vary with different camera parameters. 
Directed edges are added from every node in the graph to the event of interest. \textcolor{black}{More specifically, all neighboring events (nodes) will be connected to the newly arrived event (node or event of interest) that will be identified by the neural classifier}. It is worth noting that the graph could be of variable size, i.e. every sample might include a different number of nodes. 
A very important property of graph neural networks, is their ability to handle graphs of varying sizes, i.e. including variable number of nodes. 
This makes our approach more flexible since it facilitates operation on events arriving asynchronously at a variable rate. 

\subsubsection{\textcolor{black}{Message} passing on Graph - EventConv Layer}\label{sec:EventConv}

\begin{figure*}
\centering
 \includegraphics[width=0.94\textwidth]{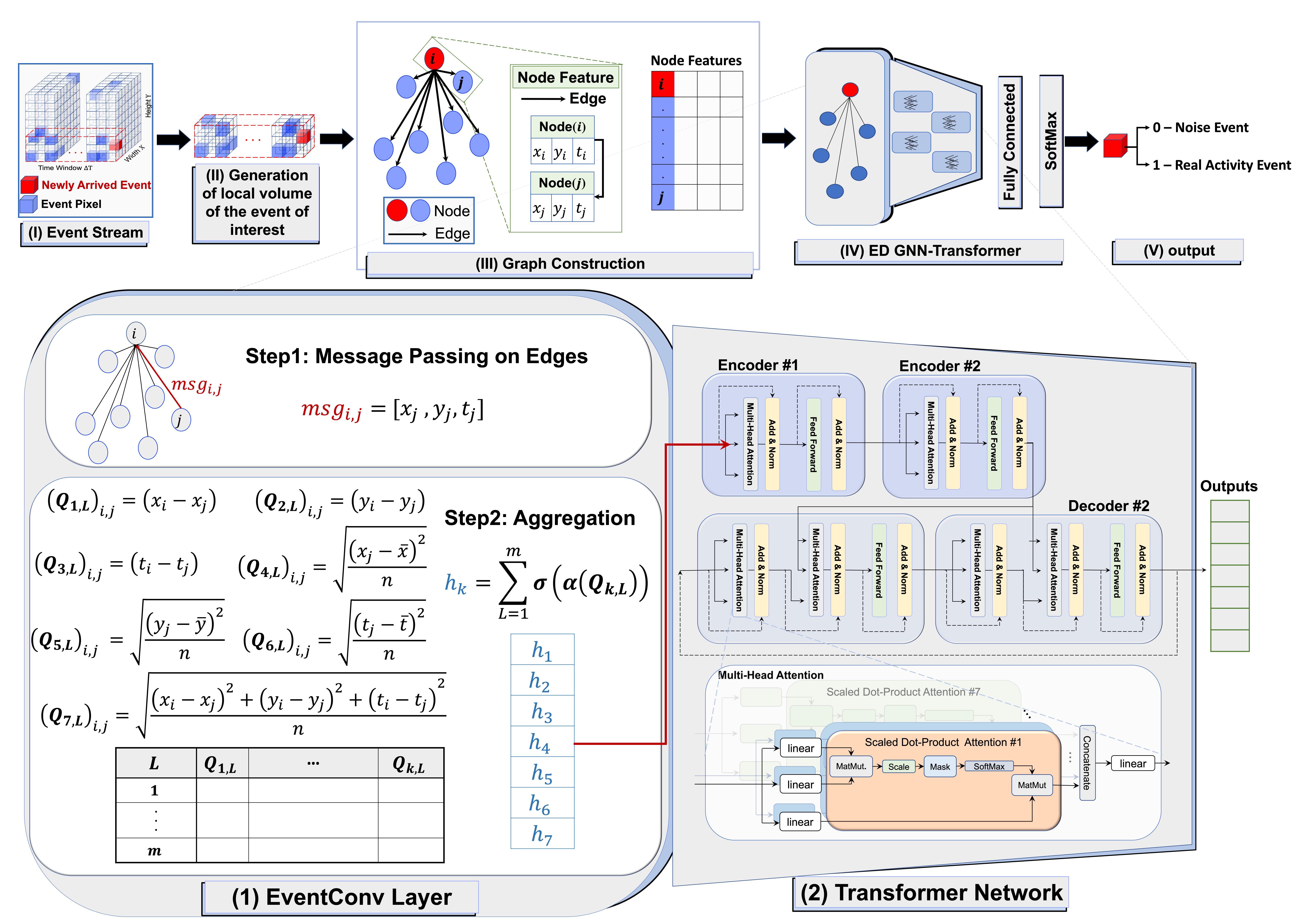}

  \caption{\textcolor{black}{Framework of our GNN-Transformer classifier for event denoising.} \textbf{Note:} \textbf{$x$} and \textbf{$y$} are the pixel coordinates at which the event occurred. \textbf{$t$} is the event's timestamp. $i$ and $j$ are the source and destination nodes where a message is transferred in Step1-(1) EventConv layer. {$Q_{1,L}, ... Q_{7,L}$} are quantities that reflect spatiotemporal properties in the graph, where $L$ represents the node index and $m$ denotes the number of events in the local volume. \textbf{$h$} is the event graph signature. \textbf{$\alpha$} is a learning parameter. \textbf{$\sigma$} is a sigmoid activation function.} 
  \label{fig:dvs_proposedApproach}
\end{figure*}

After constructing the event graph, messages are exchanged along the outgoing edges, from source nodes $j$ to the node representing the newly arrived event $i$ in the graph. 
The process of computing, sending, and aggregating the messages at the receiving node $i$ is carried out by the proposed EventConv layer. 
Every node constructs a message consisting of its three features and sends it to node $i$ for further processing. After receiving all the messages, node $i$, that  represents the newly arriving event, processes and aggregates them. 
More specifically, the average of each of the node features $<$$(x), (y), (t)$$>$ across the graph is computed (Fig. \ref{fig:dvs_proposedApproach}-(1)).
The average values $\overline{x}$, $\overline{y}$, and $\overline{t}$ are then used to estimate the spatiotemporal correlations among the events in the event graph $G$. \textcolor{black}{More specifically, the relationship between the event of interest and its neighboring events in space and time are encoded into seven quantities, which are: ($Q_{1}$) the spatial difference in x, ($Q_{2}$) the spatial difference in y, ($Q_{3}$) the temporal difference, ($Q_{4}$) the standard deviation in x, ($Q_{5}$) the standard deviation in y, ($Q_{6}$) the standard deviation in t, and ($Q_{7}$) the euclidean distance. The computations of these quantities are depicted in Fig. \ref{fig:dvs_proposedApproach}-(1) and denoted as ($Q_{1,L}, ... Q_{7,L}$), where $L$ represents the node index. These quantities were selected based on the results of an ablation study, as described in the following sections.}
Each of these quantities is passed through a linear layer followed by a sigmoid activation function prior to aggregation.
Quantities of the same type across the received messages are summed up. This operation results in a 1D vector representing a unique graph signature which is referred to as $h$. 
The uniqueness of graph $G$ signature circumvents the problem of isomorphism where two different graphs are represented by the same signature after being reduced in the aggregation stage \cite{mansour2017message}.
Message passing and aggregation steps are carried out as part of the GNN which is used in conjunction with transformers to perform classification.
The steps explained above are depicted in Fig. \ref{fig:dvs_proposedApproach}-(1).

\subsubsection{Proposed GNN-Transformer Classifier} \label{Transformer}
The overall architecture of the proposed learning-based classifier consists of two main parts including a graph neural network and a transformer. 
In this section, more details about the structure selection are explained.
\textcolor{black}{Overall, for every acquired event in the stream, a graph is constructed to reflect the spatiotemporal correlations between this event and the previous events in its neighborhood.}
The proposed GNN operates on these graphs and outputs a graph signature, previously referred to as $h$. 
This graph signature is passed to the transformer for further processing.
More particularly, the graph signature $h$ is mapped to another representation by the transformer network and finally the binary classification is performed. 
The output of the proposed GNN-Transformer is a noise-free event stream that accurately resembles the activity in the scene. 


Transformer is a sequence to sequence encoder-decoder network \cite{vaswani2017attention}. 
The self-attention mechanism encapsulates the interactions between all elements of a given sequence for structured prediction tasks. 
The attention mechanism with the Query-Key-Value (QKV) model enables the transformer to have extremely long term memory \cite{vaswani2017attention} and to execute dependencies between input and output, and consequently execute more parallelization.
The multi-head attention layer comprises multiple stacks of self-attention. 
A Multi-Head Attention mechanism encapsulates a given sequence of elements into multiple jointly complex relationships by projecting them into three learnable weight matrices, called Query, Key, and Value. 
In these matrices, computed weight distribution on the input sequence reflects the uniqueness of graph signature through assigning higher values to more representative elements. 
Basically, each element in a given input sequence in the multi-head attention layer is updated by concatenating and aggregating global representative information.

Given a graph signature $h$ with ${n}$ elements ($h_1$, $h_2$, ..., $h_n$), the objective of self-attention is to encode the global interaction information that exists among the elements. 
To achieve this, three learnable weight matrices are defined: Queries (${W}^Q \in {R}^{n\times d_q }$), Keys (${W}^K \in {R}^{n\times d_k }$), and Values (${W}^V \in {R}^{n\times d_v }$), where $W$ is the learnable weight matrix, $n$ is the size of the input features in $h$, and $d_q$, $d_k$, and $d_v$ represent the dimensions of query, key, and value vectors, respectively, $d_q=d_k=d_v=n$ in our model.
In the first step, the input sequence $h$ is projected onto these weight matrices to obtain $Q = hW^Q$, $K=hW^K$ and $V=hW^V$. 
$Z \in {R}^{n \times d_v}$ is the output of self-attention layer and is computed as follows:\useshortskip

\begin{equation}\centering
Z({Q},{K},{V})=softmax(\frac{Q{K}^T}{\sqrt{d_q}}){V}
\end{equation}\useshortskip

The most commonly used attention functions are the additive attention \cite{bahdanau2014neural} and dot product attention \cite{vaswani2017attention}. In our model, dot-product attention, which is a simple matrix multiplication, is selected to update the state within the encoder and decoder units.
This makes the attention process and its computations much faster and more space-efficient. 
In the multi-head attention process, outputs from ${d}$ self-attention units are concatenated into one vector $[{Z_1, Z_2, ..., Z_d}]$ and are then projected by an output weight matrix ${W}^o \in {R}^{nd\times n }$, as follows:\useshortskip

\begin{equation}\centering
MultiHead({Q},{K},{V})=Concat({Z_1, Z_2, ..., Z_d}){W}^o
\end{equation}

Furthermore, the multi-head attention transformer facilitates identification of jointly complex relationships and makes the model easier to interpret.

\noindent \textbf{Transformer encoder:} The architecture of the encoder and decoder layers within the transformers follows the original structure in \cite{vaswani2017attention} which consists of a multi-head self-attention unit and a feed-forward network. 
The mathematical operations in a single encoder unit can be formulated as follows:

\useshortskip
\begin{equation}\centering
q_i=k_i=v_i = \textbf{{LN}}(h_{i-1})
\end{equation}
\useshortskip
\begin{equation}\centering
y_{i-1}=h_{i-1}
\end{equation}
\useshortskip 
\begin{equation}\centering
y_i'=\textbf{MHA}(q_i, k_i, v_i) + y_{i-1}
\end{equation}
 \useshortskip 
 \begin{equation}\centering
y_i=\textbf{FFN}(\textbf{LN}(y_i’))+ y_i’\; \; \; , \;  i=1,2, ... N
\end{equation}
\useshortskip  
\begin{equation}\centering
[F_{Ei}, F_{Ei+1},...,F_{EN}]=[y_i, y_{i+1},..., y_N]
\end{equation}
\noindent where $N$ denotes the number of encoder layers, \textbf{MHA} represents the multi-head self-attention module, \textbf{LN} denotes the operation of layer normalization \cite{LN}, and \textbf{$F_E$} denotes the output of the decoder layer. 
\textbf{FFN} is the feed-forward network which contains two fully connected layers with a ReLU activation function in between as in (\ref{eq:FNN}).
\useshortskip
\begin{equation}\centering
\textbf{FFN}(x)= max(0, xW_1 + b_1)W_2 + b_2 
\label{eq:FNN}
\end{equation}
\useshortskip

\noindent \textbf{Transformer decoder:} For the Transformer decoder unit, it takes the decoder's outputs as inputs and has two multi-head self-attention modules (\textbf{MHA}) followed by a feed forward network (\textbf{FFN}). 
The mathematical operations within a single decoder unit can be formulated as follows: \useshortskip
\begin{equation}\centering
z_{i-1}=[F_{Ei}, F_{Ei+1},...,F_{El}]
\end{equation}\useshortskip \useshortskip
\useshortskip \useshortskip \useshortskip
\useshortskip
\useshortskip
\begin{equation}\centering
q_i=k_i= v_i=\textbf{LN}(z_{i-1}) 
\end{equation}\useshortskip 

\begin{equation}\centering
z_i'= \textbf{MHA}(q_i, k_i, v_i) + z_{i-1}
\end{equation}\useshortskip 

\begin{equation}\centering
q_i'=k_i'=v_i'=\textbf{LN}(z_{i-1})
\end{equation}\useshortskip 

\begin{equation}\centering
z''_{i}= \textbf{MHA}(q_i’, k_i’, v_i’) + z_{i-1}'
\end{equation} \useshortskip 
\begin{equation}\centering
z_i=\textbf{FFN}(\textbf{LN}(z''_{i}))+ z''_{i} \; \; \; , \;   i=1,2,…,l
\end{equation}\useshortskip 

\begin{equation}\centering
[F_{Di}, F_{Di+1},...,F_{D{l}}]=[z_i, z_{i+1},..., z_{l}]
\end{equation}\useshortskip
\noindent where $l$ denotes number of decoder layers and $F_{D}$ represents the output of the transformer unit ($F_{D} \in {R}^{n \times 1}$) which reveals important features to uniquely represent the graph signature ($h$).   

The output of the coupled GNN-Transformer is finally passed to a fully connected layer that generates a $2 \times 1$ tensor for every sample in the dataset, where 2 is the number of classes: real log-intensity change or noise. The output tensor is passed to a softmax function (\ref{eq:softmax}) , where it is rescaled so that the elements are in the range $[0,1]$ and sum up to unity. The rescaled elements represent the probabilities that the event under investigation represents noise or real-activity, respectively.

\begin{equation}\centering
\textbf{Softmax}(x_i)=  \frac{e^{x_i}} {\sum_{j=1}^2 e^{x_j}} 
\label{eq:softmax}
\end{equation}
\useshortskip 

Supervised learning is performed using the backpropagation algorithm to train the GNN-Transformer network. 
Pytorch \cite{pytorch} \textcolor{black}{implementation} is used for constructing all the neural networks and performing training and testing. 
The training process is carried out to minimize the cross-entropy loss function using the Adam optimizer \cite{kingma2014adam} with a learning rate of 0.001.

\noindent\textbf{\textit{Ablation Study:}} \textcolor{black}{To select the most suited structures of both the GNN and the transformer, an automated search routine was developed. The automated search routine spanned several parameters including the graph structure, the message operation, the aggregation functions, the number of EventConv layers in the GNN, the activation functions, and the number of encoder-decoder units in the transformer. Such parameters reflect the nonlinear capacity of the model and hence need to be carefully selected to best suit the problem in question. It was observed that several architectures have achieved comparable performance and were able to correctly classify the majority of real-activity and noise events.}

\begin{figure*}
\centering
 \includegraphics[width=0.82\textwidth]{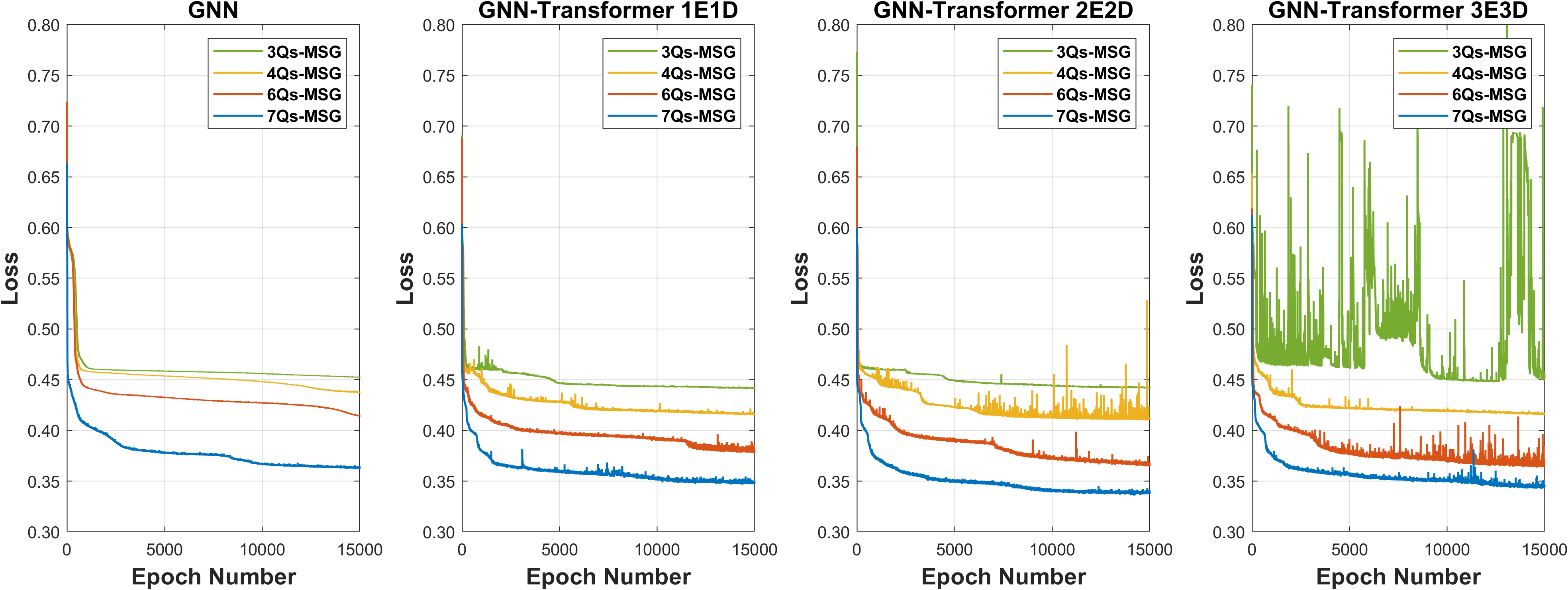}

  \caption{\textcolor{black}{Ablation study results - loss curves obtained upon training various network architectures as part of the automated search for the best suited neural network architecture.}}
  \label{fig:loss}
\end{figure*}

\textcolor{black}{
Figure \ref{fig:loss} reports the loss obtained by the highest performing architectures on the training dataset among the tested neural networks. The loss curves are grouped based on the adopted neural network architecture; GNN, GNN in conjunction with a transformer of a single encoder-decoder layer (GNN-Transformer 1E1D), GNN in conjunction with a transformer of a double encoder-decoder layer (GNN-Transformer 2E2D), and GNN in conjunction with a transformer of a triple encode-decoder layer (GNN-Transformer 3E3D). For every architecture, the number of quantities composing the messages that characterize the spatiotemporal correlation within the graph was varied. More specifically, four combinations of quantities in the message were tested as indicated below: }

\begin{itemize}
\item \textcolor{black}{\textbf{3Qs-MSG:} $Q_1$, $Q_2$, $Q_3$}
\item \textcolor{black}{\textbf{4Qs-MSG:} $Q1$, $Q2$, $Q3$, $Q7$}
\item \textcolor{black}{\textbf{6Qs-MSG:} $Q1$, $Q2$, $Q3$, $Q4$, $Q5$, $Q6$ }
\item \textcolor{black}{\textbf{7Qs-MSG:} $Q1$, $Q2$, $Q3$, $Q4$, $Q5$, $Q6$, $Q7$}
\end{itemize}

\textcolor{black}{The performance of all the attempted networks is evaluated using unseen testing datasets, which are composed of streams of events obtained experimentally. The performance evaluation metrics used to compare the training and validation results are the accuracy, signal ratio, noise ratio, and signal to noise ratio as computed with respect to the ground truth labels obtained using our proposed KoGTL for each event. }

\textcolor{black}{
Training and testing results have proven that the GNN-Transformer architecture with 7Qs-MSG in the EventConv layer as described in Section \ref{sec:EventConv} and a transformer with a double encoder-decoder layer showed the best performance among all candidate neural classifiers in terms of the noise filtration accuracy as reported in Table \ref{train2test2_2} in the supplementary material. The proposed GNN-Transformer architecture is depicted in Fig. \ref{fig:dvs_proposedApproach}-IV.}

\textcolor{black}{It is worth noting that the quantities included in the messages play a pivotal role in reflecting the spatiotemporal correlation of the event and its neighboring events and thus in the overall performance of the filter as clearly shown in loss curves of the GNN-Transformer 3E3D. More specifically, although the architecture of the neural network was complex enough, the number of quantities in the message drastically affected the filter's performance.}

\setlength{\belowcaptionskip}{-15pt}

\section{Experimental Evaluations}
\label{sec:results}

The proposed GNN-Transformer algorithm  for event denoising is tested qualitatively and quantitatively in multiple scenarios to demonstrate its validity, effectiveness and generalization.
The training process including training and testing data preparation is described in Section \ref{dataset}.
In Section \ref{EM}, the evaluation metrics used to quantify the results are presented.
Section \ref{seq:quantitative} presents the quantitative performance analyses of the developed GNN-Transformer model. 
Moreover, the GNN-Transformer model is benchmarked against other existing event denoising methods, where the developed model's capability, effectiveness, and validity are discussed. 
In addition, the performance of the model is evaluated qualitatively on part of the datasets that we have recorded, but have not exposed to the network during training, as well as several publicly available datasets as presented in Section \ref{seq:qualitative}. This is to prove the model's generality and robustness to various illumination conditions and unseen data.

\subsection{Training and Testing Datasets} \label{dataset}

Training and testing datasets are constructed from experiments recorded in our lab as well as other publicly available datasets. Training is exclusively done using our recorded dataset because of the availability of ground truth labels to support supervised learning. Testing, on the other hand, is done on both recorded and publicly available datasets where quantitative and qualitative evaluations are done. 

Recorded experiments were conducted following the approach described in Section \ref{Sec:KoGTL} using the iniVation’s DAVIS346C dynamic vision sensor \cite{davis346c}. Four lighting conditions were used to record experiments; very good lighting ($\thicksim$750lux), office lighting ($\thicksim$300lux), low light condition ($\thicksim$5lux), and Moon light condition ($\thicksim$0.15lux). Every experimental scenario includes scenes recorded when the camera is static or is starting translational motion, and scenes recorded when the camera is moving in four different directions. In the former case, static noise pixels can be detected and learned accordingly. The latter cases exhibit the dynamic nonlinear nature of event and noise generation as well as spatiotemporal correlations of an event and its neighborhood when the camera is in motion.

Samples from the experiments recorded under very good lighting ($\thicksim$750lux) and low light condition ($\thicksim$5lux) were used for quantitative analysis (training and testing). \textcolor{black}{Each sample consists of a newly arrived event and its corresponding neighboring events within the defined spatial and temporal window.} More specifically, for each scenario, a total of 8000 samples were randomly selected from each of the five scenes (static and motion in four directions); 4000 real-activity events and 4000 noise samples. This is to ensure that the training dataset is balanced and is not biased towards one class more than the other. Hence, a total of 80k samples constitute the dataset; 80\% of which are used for training and 20\% are used for testing. 

Moreover, qualitative analysis of the model's performance on two recorded experiments ($\thicksim$300lux and $\thicksim$0.15lux) and \textcolor{black}{eleven publicly available datasets was carried out. The publicly available datasets \cite{publicDataset} include indoor and outdoor scenarios and were recorded at numerous illumination conditions and using different motion dynamics as summarized in Table \ref{tabledataset}.}
\vspace{+4mm}
\begin{table}[hb!]
\centering

\caption{\textcolor{black}{Description of the publicly available datasets used from \cite{publicDataset}}}
\label{tabledataset}
\begin{adjustbox}{width=0.48\textwidth}
\setlength{\arrayrulewidth}{0.5pt}

\begin{tabular}{ r l  c } 
\hline
\textbf{\Large{Name} }&\Large \centering \textbf{Scene Description} & \large \begin{tabular}[c]{@{}c@{}}\Large \textbf{Light}\\\large \textbf{Condition} \\\large\textbf{(lux)}\end{tabular}
\\ \hline
\\ & & \\
\Large\textbf{\textit{Simple}-Scene} & \Large \begin{tabular}[c]{@{}l@{}}Simple 6DOF camera motions looking at simple\\ objects and scenes with vibrant colors.\end{tabular} & 
\\ & &  \\
\Large{\textit{SimpleFruit} }& \Large \begin{tabular}[c]{@{}l@{}}Colorful fruits, fluorescent and window lighting. \end{tabular}& \Large 1000 \\
& & \\\Large{\textit{SimpleObjects} }& \Large 
\begin{tabular}[c]{@{}l@{}}\\ Colorful everyday objects, fluorescent and \\window lighting. \end{tabular} 
& \Large 1000 
\\
\Large{\textit{SimpleObjectsDynamic}} &  \Large \begin{tabular}[c]{@{}l@{}}\\ Colorful everyday objects being picked up, \\fluorescent and window lighting. \\\end{tabular}  &\Large  1000 \\& & \\
\Large{\textit{SimpleWires1}} & \Large  \begin{tabular}[c]{@{}l@{}}Colorful rolls of wire, fluorescent and window \\lighting.\end{tabular}  
& \Large 400 \\ 

\\\hline
\\ & & \\
\Large\textbf{\textit{Indoors}-Scene}&\Large  \begin{tabular}[c]{@{}l@{}}Natural indoor scenes including office, kitchen, \\rooms and corridors.\end{tabular}&\\ & & \\
\Large{\textit{IndoorsCorridor} } & \Large \begin{tabular}[c]{@{}l@{}}Walking down dimly lit corridor, into room \\with bright windows.\end{tabular} & \Large 80-1000\\ & & \\
\Large{\textit{IndoorsDark25ms} }  &\Large  \begin{tabular}[c]{@{}l@{}}Desk illuminated by two monitors, exposure\\ set to 25ms. \end{tabular} 
& \Large 2  \\ 
\Large{\textit{IndoorsFootball1} }  & \Large Foosball table, fluorescent
  lighting. & \Large 200   \\& & \\
\Large{\textit{IndoorsKitchen1} }   & \Large People in kitchen, fluorescent
  lighting.  & \Large 200  \\ \\
\hline \\ & & \\
 \Large\textbf{\textit{Driving}-Scene}  & \Large \begin{tabular}[c]{@{}l@{}} Footage from front windshield of car driving \\around country, suburban and city landscapes.\\ Features tunnels, traffic lights,vehicles and\\ pedestrians during the day in sunny conditions.\end{tabular}  &                                                      \\ & & \\
  \Large{\textit{DrivingCity4}} &  \Large \begin{tabular}[c]{@{}l@{}}Driving around the city,features tunnel and\\ light traffic.\end{tabular} 
  & \Large 200-100,000         \\ & & \\

\Large{\textit{DrivingTunnel}}&\Large \begin{tabular}[c]{@{}l@{}}Driving into long
  tunnel (15 seconds) and \\out into bright sunlight. \\\end{tabular} &\Large 200-100,000 \\ 
& & \\
\Large{\textit{DrivingTunnelSun}}              & \Large \begin{tabular}[c]{@{}l@{}}10 second tunnel followed by direct sun in\\ field of view.\end{tabular}  
&\Large 200-100,000  \\ 
\hline
\end{tabular}
\end{adjustbox}

\end{table}

Prior to training the model, every sample event and its corresponding neighborhood are used to construct a graph, which is used as the input to the graph neural network. The size of the neighborhood, i.e. the local volume, is selected to be a maximum of 10 nodes (or events) within 5 by 5 pixels window centered at the event of interest in the preceeding 50 ms. In case more events were acquired in this volume, only the latest 10 are included in the graph. 
It is worth mentioning that the volume size was selected after several experiments with varying volume parameters. It was observed that 10 neighboring events in the local volume are sufficient to delineate the spatiotemporal correlations and hence make a decision on whether the event of interest is real or noise.

To expedite training and convergence, it is common practice to normalize all the inputs to the neural network to a common range. In this work, all inputs are rescaled to the range $[0.05, 0.95]$, excluding values very close to 0 and 1 to avoid the issue of neuron saturation which causes the problem of vanishing gradients. For example, the minimum and maximum values of sigmoid are 0 and 1 respectively. The corresponding derivative at those values drops to zero, causing gradients to vanish. 


\subsection{Evaluation Metrics}\label{EM}

To quantitatively evaluate the performance of the proposed denoising model and compare to state-of-the-art models on training and testing datasets, four evaluation metrics are used: $Accuracy$, Signal Ratio  (\textit{\text{SR}}), Noise Ratio (\textit{\text{NR}}), and Signal to Noise Ratio (\textit{\text{SNR}}).

\paragraph{Accuracy}
This metric measures the model's ability to correctly predict real activity events and noise, as defined in (\ref{accu}). \useshortskip 
\begin{equation}\centering \label{accu}
Accuracy= \frac {\text{TP}+{\text{TN}}}{\text{TP}+\text{TN}+ \text{FP}+\text{FN}}
\end{equation}
where \text{TP}, \text{FP}, \text{TN}, and \text{FN} are the number of true positives, false positives, true negatives and false negatives pixels, respectively. \text{TP} indicates the number of events that are correctly predicted as real activity events, whereas \text{TN} indicates the number of events that are correctly predicted as noise.

\paragraph{Signal Ratio (\textit{\text{SR}})}
This metric represents the proportion of correctly predicted real-activity events with respect to the total number of real-activity events in the scene, \textcolor{black}{which is also known as precision}, as defined in (\ref{eq:SR}). 

\begin{equation}\centering \label{eq:SR}
\textit{\text{SR}}= \frac {\text{TP}}{\text{TP}+\text{FP}}
\end{equation}


\paragraph{Noise Ratio (\textit{\text{NR}})}
This metric represents the proportion of incorrectly predicted noise events with respect to the total number of noise events in the scene, \textcolor{black}{which is also known as the false omission rate}, as defined in (\ref{eq:NR}). 

\begin{equation}\centering \label{eq:NR}
\textit{\text{NR}}= \frac {\text{FN}}{\text{TN}+\text{FN}}
\end{equation}

\paragraph{Signal to Noise Ratio (\textit{\text{SNR}})}
This metric is the ratio of the number of correctly predicted real-activity events to the number of noise events incorrectly labeled as real-activity events as described in (\ref{eq:SNR}). 

\begin{equation}\centering \label{eq:SNR}
\textit{\text{SNR}}= \frac {\text{TP}}{\text{FN}}
\end{equation}

The performance of the denoising model is considered better with higher \textit{\text{SR}} and \textit{\text{SNR}} values and lower \textit{\text{NR}} values.

    \vspace{-4mm}

\subsection{Quantitative Results} \label{seq:quantitative}

\subsubsection{Evaluation on Training and Testing Datasets} 
In this section, the performance of the proposed GNN-Transformer based Event Denoising model is compared against state-of-the-art denoising methods, namely \textcolor{black}{EDnCNN \cite{D9_Baldwin2020a}}, Yang Filter \cite{D6_Feng2020}, Khodamoradi Filter \cite{D7_Khodamoradi2017}, Liu Filters \cite{D1_Liu2015}, and Nearest Neighbor NNb filter \cite{D2_Padala2018}. All filters are tested on the same dataset, which was used to train our proposed approach. The dataset was randomly split into training and testing subsets, where 80\% of the samples were used for training and 20\% were used for testing (not exposed to the network during training).

\textcolor{black}{EDnCNN filter's parameters were set to those mentioned in their published trained model which consists of $3\times3$ convolutional layers followed by two fully connected layers. To filter an event, a spatiotemporal window of $25\times25\times5s$ centered at that event pixel is considered to construct the input feature to the model. More specifically, a $25\times25\times k\times2$ matrix is populated with the $k$ most recent positive and negative events that were received prior to the event of interest, where $k$ was set to 2. The pre-trained EDnCNN model parameters \cite{D9_Baldwin2020a} were used to perform accuracy evaluations on both of our training and testing datasets.} Yang filter's parameters were set to the default values reported in \cite{D6_Feng2020}. More specifically, the time window was set to 5ms, spatial window is 5 by 5 pixels, and the density is 3. As for Khodamoradi filter, the time window was set to 1ms, as in \cite{D7_Khodamoradi2017} and \cite{D6_Feng2020}. Two down-sampling factors $S$ of Liu’s filter were used $S$= 1,2 where the timestamp of 2$\times$2 and 4$\times$4 pixels were stored in one memory cell and the time window was set to 1ms, as test in \cite{D6_Feng2020}. The working principle of Liu and Khodamoradi filters was previously mentioned in Section \ref{sec:related_work}, Fig. \ref{fig:BAmethods}b and Fig. \ref{fig:BAmethods}c, respectively. 
Lastly, for Nearest Neighbor NNb filter, the size of the event's local volume is set to 3 by 3 pixels for 1ms, as reported in their work \cite{D2_Padala2018}. The performance of these denoising methods was  compared to that of the proposed GNN-Transformer approach as presented next.

\textcolor{black}{Table \ref{train2test2} reports the filtration accuracy achieved by the GNN-Transformer network, EDnCNN filter, Yang filter, Khodamoradi filter, Liu filter and NNb filter when evaluated on the training and testing datasets. It is worth mentioning that the training and testing datasets have equal numbers of real and noise events (50\% real events and 50\% noise events). It is observed that the GNN-Transformer outperforms all the other alternatives in terms of filtration accuracy. The proposed model has outperformed EDnCNN by $10.6\%$ on the training dataset and $8.4\%$ on the testing dataset. Is has also achieved $12\%$ higher training and testing accuracy compared to Yang filter. While Yang filter has shown the best performance compared to other conventional filters (Khodamoradi, Liu, and NNb filters) in terms of filtration accuracy.} 

A high \textit{\text{SNR}} value does not necessarily mean that a filter's performance is better than others. Rather, a high \textit{\text{SNR}} value, a high \textit{\text{SR}} value, and a low \textit{\text{NR}} value together would indicate a good filtering performance. A clear example is the Khodamoradi filter, which achieved the highest \textit{\text{SR}} (99\%) and the highest \textit{\text{NR}} (92\%) values among other filters. These values mean that all input data have been considered real-activity and no noise filtration took place. In other words, the filter could not distinguish between the incoming real-activity events and the accompanying noise. 

Another example is Liu’s filter, which achieved the lowest \textit{\text{NR}} (1-2\%) and a relatively low \textit{\text{SR}} (10-30\%). In this case, most of the input data have been considered as noise. This implies the weak denoising capability of Liu's filter. Meaningful real-activity events have been filtered out and consequently scene perception algorithms would fail to operate as expected.

To conclude, the best event denoising model is expected to have a high accuracy, \textit{\text{SR}}, and \textit{\text{SNR}}, and a low \textit{\text{NR}}. Thus, our proposed GNN-Transformer has clearly outperformed all alternative filters and proved its capability to generalize to unseen datasets. Table \ref{train2test2} compares the number of correctly and incorrectly predicted real-activity events from the training and testing datasets.  
\vspace{+4mm}

\begin{table}[hbt!] \centering

\caption{ Performance of the GNN-Transformer classifier compared to state-of-the art denoising methods on the training and testing datasets}
\label{train2test2}
\begin{adjustbox}{width=0.49\textwidth, height=0.11\textheight}
\centering
\begin{tabular}{!{\VRule[1.25pt]}c|cccc|c!{\VRule[1.25pt]}}

\specialrule{1.25pt}{0pt}{0pt}
\multicolumn{6}{!{\VRule[1.25pt]}c!{\VRule[1.25pt]}}{\centering{Training Dataset}}

\\ \hline

 Event Denoising Method &
  TP 
&  FP
& TN
 & FN &
Filtration $Accuracy$ 
\\ \hline
Yang Filter \cite{D6_Feng2020}&
15529&
16471&
29012&
2988&
69.60\%
\\
Khodamoradi Filter \cite{D7_Khodamoradi2017}&
31889&
111&
2526&
29474
&
53.77\% 
\\
Liu Filter \cite{D1_Liu2015} (SubGroup by 2) &
3665&
28335&
31225&
775 &54.52\% 

\\
Liu Filter \cite{D1_Liu2015} (SubGroup by 4)&
10149&
21851&
28429&
3571 &
60.28\% 
\\
NNb Filter \cite{D2_Padala2018} &
7594&
24406&
30313&
1687&
59.23\% 
\\\textcolor{black}{EdnCNN \cite{D9_Baldwin2020a}}
&\textcolor{black}{18830}
&\textcolor{black}{13170}
&\textcolor{black}{ 27082}
&\textcolor{black}{4918}
& \textcolor{black}{71.73\%}

\\ \hline
GNN-Transformer (ours)&
27012&
4988&
25684&
6316 &
{\textbf{82.34\%}}
\\



\specialrule{1.25pt}{0pt}{0pt}

\multicolumn{6}{!{\VRule[1.25pt]}c!{\VRule[1.25pt]}}{\centering{Testing Dataset}}

\\ \hline

 Event Denoising Method &
  TP 
&  FP
& TN
 & FN
 &
Filtration $Accuracy$
\\ \hline
Yang Filter \cite{D6_Feng2020}&
3831&
4169&
7220&
780&
69.07\% 

\\
Khodamoradi Filter \cite{D7_Khodamoradi2017}&
7977&
23&
670&
7330

&
54.04\% 
\\
Liu Filter \cite{D1_Liu2015} (SubGroup by 2) &
925&
7075&
7829&
171
&
54.71\% 

\\
Liu Filter \cite{D1_Liu2015} (SubGroup by 4)&
2451&
5549&
7092&
908
&
59.64\% 
\\
NNb Filter \cite{D2_Padala2018} &
1889&
6111&
7564&
436
&
59.08\% 
\\\textcolor{black}{EdnCNN \cite{D9_Baldwin2020a} }&
\textcolor{black}{4722}&
\textcolor{black}{3278}&
\textcolor{black}{6790}&
\textcolor{black}{1210}
&
\textcolor{black}{71.95\%}

\\ \hline
GNN-Transformer (ours)&
6403&
1597&
6513&
1487&\textbf{80.73\%} 
\\
  \specialrule{1.25pt}{0pt}{0pt}
      
\end{tabular}

\end{adjustbox}

\end{table}

\begin{figure*}[]
\centering
  \includegraphics[width=0.615\textwidth]{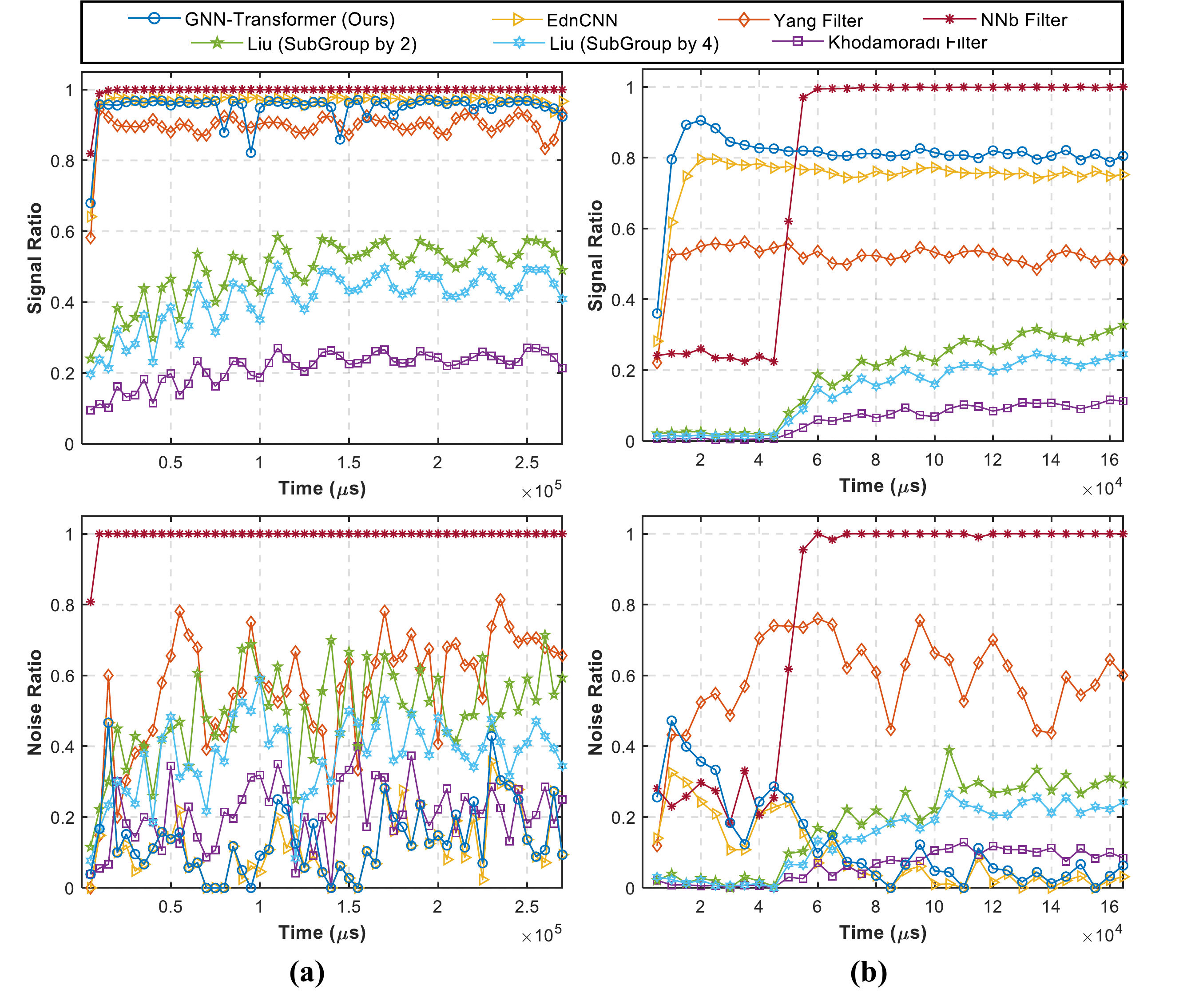}
  \caption{Signal Ratio (\textit{SR}), Noise Ratio (\textit{NR}), and Signal to Noise Ratio (\textit{SNR}) event denoising performances of the GNN-Transformer Model and state-of-the-art denoising methods - using sample stream of events recorded (a) at $\thicksim$750lux (b) at $\thicksim$5lux. }
  \label{fig:period}
\end{figure*}

\begin{figure*}
\centering
 \includegraphics[width=0.659\textwidth]{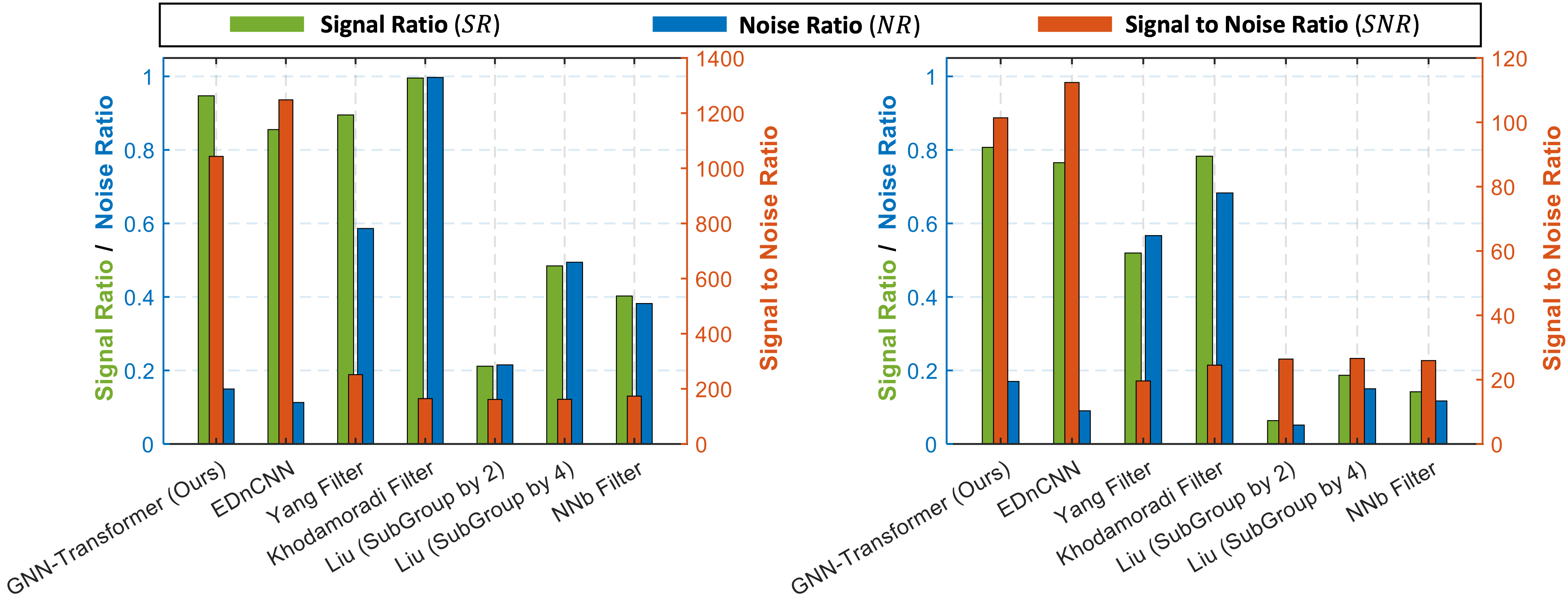}
  \caption{Signal Ratio (\textit{SR}), Noise Ratio (\textit{NR}), and Signal to Noise Ratio (\textit{SNR}) event denoising performances of the GNN-Transformer model and state-of-the-art denoising methods - using sample stream of events recorded (a) at $\thicksim$750lux (b) at $\thicksim$5lux.  \textcolor{black}{\textcolor{black}{The performance of the denoising model is considered better with higher \textit{\text{SR}} and \textit{\text{SNR}} values and lower than \textit{\text{NR}} values. It can be observed that the best performing denoising methods are ours and EDnCNN \cite{D9_Baldwin2020a}. However, for fair comparison and for these results to make sense, the metrics have to be analyzed collectively. It was observed that EDnCNN has considered a large number of events as noise, which decreased the \textit{NR} value compared to ours. However, a significant amount of these filtered events belongs to meaningful features, i.e. was incorrectly labeled as noise, which resulted in a lower \textit{SR} value than ours.}}}
  \label{fig:bar}
\end{figure*}
\setlength{\textfloatsep}{1pt}

\subsubsection{Evaluation on our Recorded Dataset - Continuous Stream of Events}
In this section, the proposed model is tested online on a continuous stream of events then compared to state-of-the-art \textcolor{black}{denoising} techniques. In other words, instead of randomly selecting samples from the recorded experiments, the full stream of events generated by DVS is passed through each filter, which is then evaluated, as per our labeled dataset.  

Filtering techniques were tested in two scenarios; the experiments recorded at $\thicksim$750lux and $\thicksim$5lux. In the first scenario, filtering was done over 600ms, where \textit{\text{SR}} and \textit{\text{NR}} were evaluated every 10ms as shown in Fig. \ref{fig:period}a. The second scenario was run for 170ms and evaluation was done at 5ms intervals as shown in Fig. \ref{fig:period}b. Evaluations of \textit{\text{SR}}, \textit{\text{NR}}, and \textit{\text{SNR}} over the full period of time for both scenarios are depicted in Fig. \ref{fig:bar}a and Fig. \ref{fig:bar}b. The total number of events included in this test is 7M and 0.1M for the first and second scenarios, respectively.

\textcolor{black}{It is evident, through the conducted tests, that our proposed GNN-Transformer based event denoising technique has achieved the best filtering performance compared to all the other filters. This proves the effectiveness of the proposed event denoising approach and shows robustness to different camera motion dynamics under illumination variations. According to our evaluations, the second-best learning-based event-denoising technique is the EDnCNN \cite{D9_Baldwin2020a} filter and the best conventional event-denoising filter is Yang filter \cite{D6_Feng2020}. Thus, further qualitative performance assessments of our proposed approach are conducted against those two filters only as presented in Section \ref{seq:qualitative}. }

\subsubsection{Computational Time Complexity and Memory Analysis}

In this section, time and memory analyses of the proposed approach will be discussed and compared to EDnCNN filter since both are based on using neural networks. A set of 10,000 event samples was selected from the \textbf{\textit{stairs}} dataset presented in \cite{D9_Baldwin2020a} to conduct the timing analysis.

\textcolor{black}{The computational time analysis of the proposed algorithm was carried out on an ASUS laptop with Intel core $i7-7700HQ @ 2.80 GHz \times 4$, NIVIDIA GeForce GTX 1050 Ti 4GB. The analysis was done with and without GPU support in two modes; \textit{Sequential mode:} events were passed to the filter successively, one after the other, and \textit{Batch mode:} all events were passed to the filter as a single batch. 
The time needed to filter the events in each mode was recorded for both filters as listed in Table \ref{fig:comput}. In all cases, the time needed to complete the filtration was shorter using our proposed approach compared to EDnCNN. However, our approach achieved a large speed-up of up to two orders of magnitude in the \textit{batch} mode compared to the other filter when run on CPU, and a speed-up of up to one order of magnitude when run on GPU. This speed-up is significant as operation in \textit{batch} mode is certainly necessary due to the high temporal resolution of the event camera, and due to the working principle of the event camera that enables $346 \times 260$ pixels to be active simultaneously. In other words, the proposed approach is capable of handling batches of events concurrently in a very short period of time, and hence preserves the high temporal resolution of the sensor. It is also worth noting that the proposed approach exhibited the fastest performance when processing events in a batch mode on a CPU, which obviates the need for sophisticated hardware to achieve fast and accurate noise filtration. This makes the proposed approach suitable for limited computational power and resource-constrained platforms such as high speed UAV control \cite{2_uav_control}, UAV navigation \cite{1_navigation}, and space applications \cite{salah2022}. }

\textcolor{black}{ To project this analysis on a real-world scenario, consider the application of autonomous car driving where neuromorphic vision could be employed to observe the environment during navigation. As the speed of the vehicle increases, the number of generated events will proportionally increase resulting in a tremendous amount of events for processing. Faster processing of visual observations will thus result in a faster response to changes in the vehicle's surroundings. This will definitely reduce the probability of collisions and will enhance the effectiveness of the overall system. }



The overall memory requirement per event classification is $5\times 5\times N_g$, where $N_g$ is the number of events per graph and could range from 1-10 events. Whereas in EDnCNN, the size of the input feature is $25\times25\times2\times2$. This clearly shows that our approach is more memory efficient than EDnCNN, where in case the graph in our approach had 10 nodes (which is the maximum number of nodes per graph), the memory requirements are 10 times less than that of EDnCNN.

\setlength{\belowcaptionskip}{5pt}

\begin{table}[hbt]
 \caption{\textcolor{black}{Time in seconds to filter events using our proposed approach and EDnCNN method \cite{D9_Baldwin2020a}. \textbf{Note} that $\mu$ and $\sigma$ represent the mean and standard deviation, respectively.}}
  \label{fig:comput}

\begin{adjustbox}{width=0.49\textwidth}

\begin{tabular}{|c|c|c|c|}
\hline
          \multirow{2}{*}{Processing Unit}&            \multirow{2}{*}{Event Denoising Model}    & Sequential-mode & Batch-mode \\
                           & &  $\mu$ $\pm$ $\sigma$ ($sec$)              &  $\mu$ ($sec$)    \\\hline

 \multirow{2}{*}{CPU}& \ EDnCNN \cite{D9_Baldwin2020a} &\num{1.46e-2} $\pm$  \num{2.38e-3} 

                           & \num{1.54e-3}
    \\

          &             \textbf{ GNN-Transformer} & \textbf{\num{1.28e-3} $\pm$  \num{2.07e-4}  }           & \textbf{\num{5.24e-5}  }  \\\hline
\multirow{2}{*}{GPU}&
 EDnCNN \cite{D9_Baldwin2020a}&  \num{7.19e-3} $\pm$  \num{3.12e-3} 

             & \num{3.01e-04}
    \\

                  &    \textbf{GNN-Transformer}   & \num{1.69e-3} $\pm$ \num{2.65e-4 }           & \num{6.77e-5}
 \\\hline
\end{tabular}
\end{adjustbox}
\end{table}

\setlength{\belowcaptionskip}{5pt}

\setlength{\belowcaptionskip}{-5pt}
    \vspace{-4mm}

\subsection{Qualitative Results} \label{seq:qualitative}

\begin{figure}
\centering
 \includegraphics[width=0.333\textwidth]{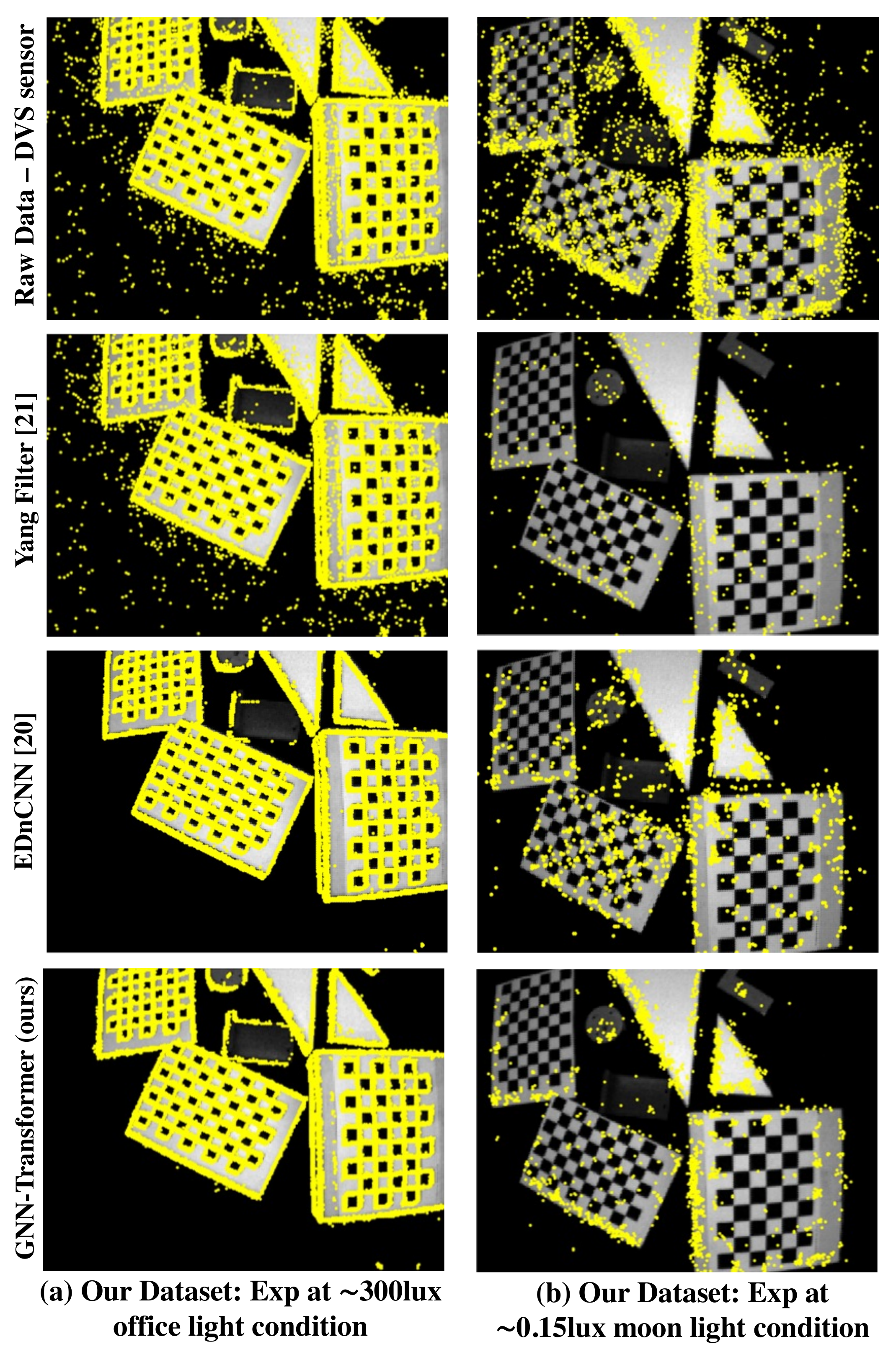}
\caption{Denoising results tested on our Dataset (unseen data), denoised events from DVS (yellow dots) overlaid on corresponding APS image. 
} \label{qual_1}
\end{figure}

In this section, two experiments from our recorded dataset, particularly those recorded at $\thicksim$300lux and $\thicksim$0.15lux, are used to qualitatively analyze the denoising performance of the proposed model against and \textcolor{black}{EDnCNN and Yang filters}. Sample filtering results, superimposed on APS images for better visualization, are depicted in Fig. \ref{qual_1}. The results clearly show that our model has filtered out most of the background activity noise and maintained events representing relative motion of meaningful features in the scene as in Fig. \ref{qual_1}a. Although more scattered noise is present under low lighting conditions as shown in Fig. \ref{qual_1}b, our proposed model was able to preserve the events that represent meaningful features (edges) in the scene. Conversely, Yang filter has eliminated the majority of real-activity events from the scene, while leaving some scattered ones that could be hard to interpret as edges or meaningful features. This proves the robustness of our model against illumination variations. 

\begin{figure}[]
\centering
 \includegraphics[width=0.33\textwidth]{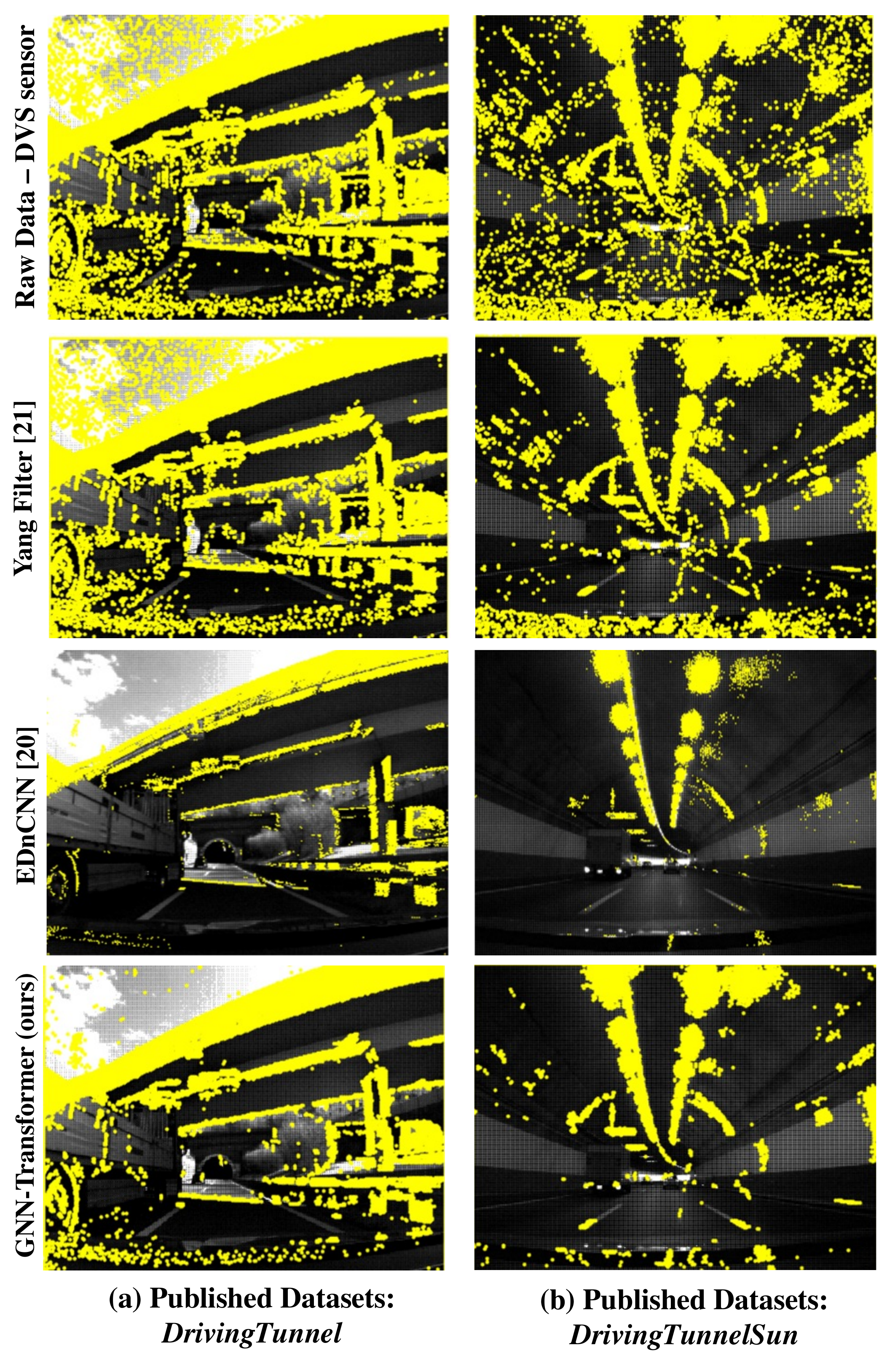}
\caption{\textcolor{black}{Sample of denoising results tested on published datasets (unseen data), denoised events from DVS (yellow dots) overlaid on corresponding APS image.}}\label{qual_2}
\end{figure}
\setlength{\textfloatsep}{7pt}

To further prove the validity and generalization of our proposed model, 
we have extensively tested it and compared it against others using \textcolor{black}{eleven} publicly available datasets. These recorded data were acquired from different camera motion dynamics (type of motion and speed) and under different lighting conditions. Fig. \ref{qual_2} shows two examples of denoised events obtained using the proposed model, \textcolor{black}{EDnCNN, and Yang filter. It was noticed that EDnCNN eliminated a large amount of events that belong to meaningful features in the scene. For instance, the filtered event stream corresponding to the scene taken from the \textit{DrivingTunnelSun} dataset shown in Fig. \ref{qual_2}a lacks significant events that represent clear intensity variations as per the corresponding APS images. Such events were classified as noise using the EDnCNN filter. The same observation can be seen in the scenes from the other datasets such as \textit{DrivingCity4} in the same figure.} Yang filter passes the majority of the events (both real and noise signals), thus making it more difficult to identify objects (edges) in the scene compared to our proposed model. Therefore, the GNN-Transformer based event denoising model generalizes well to new scenarios under various illumination conditions without any further tuning of its parameters. More results are demonstrated in the supplementary material (Appendix: Fig. \ref{qual_3}), additional results document in $<$\url{https://github.com/Yusra-alkendi/ED-KoGTL}$>$ and \textcolor{black}{video $<$\url{https://youtu.be/ZM76UaxbuJE}$>$}, which visualize the denoising performance of GNN-Transformer classifier compared to Yang Filter \cite{D6_Feng2020} and EDnCNN \cite{D9_Baldwin2020a}.

    \vspace{-2mm}

\section{Conclusion} \label{sec:conclusion} 
In this work, we developed a novel algorithm to filter out the noise associated with event streams acquired by dynamic vision sensors.
The GNN-Transformer based event-denoising algorithm exploits the spatiotemporal correlations between events in a particular neighborhood to decide whether an incoming event represents noise or a log-intensity variation in the observed scene. To train the proposed GNN-Transformer model, a novel offline event labeling technique, KoGTL, is proposed to distinguish between noise and real events in event streams recorded under challenging lighting conditions. 
The labeled DVS data is made available to the public research community for benchmarking purposes. The proposed algorithm successfully operates on event streams irrespective of camera parameters, illumination conditions, and motion dynamics.
This is attributed to the fact that the adopted graph structure of the input data preserves the spatiotemporal correlation between the events, rather than the raw properties of the events, solely. 
Such operation is carried out in the proposed EventConv layer. 
The proposed algorithm also operates on event graphs of variable sizes and thus handles the asynchronous nature of event streams.

Through extensive training and testing, the proposed algorithm has proven to achieve significantly high denoising performance under challenging illumination conditions.
Our model is also tested on \textcolor{black}{eleven} publicly available datasets which were not exposed to the network during training. 
The model is able to successfully denoise the event streams, despite the fact that the data is recorded under conditions different than those of the training data, including different environmental conditions, various camera motions, and camera parameters.
\textcolor{black}{The quantitative results have demonstrated the denoising capability of the proposed algorithm with at least 8.8\% higher filtration accuracy on testing sets compared to existing methods.}
Qualitatively, the results achieved by the proposed model have verified its effectiveness and generalization to previously unseen event graph data, irrespective to their sizes. 
This work has unveiled the power and potential of graph neural networks and transformers on event cameras.

\textcolor{black}{In the future, we plan to demonstrate the significance of our proposed denoising approach by integrating it into other event-based computer vision algorithms such as motion segmentation, object detection, object tracking, and object recognition, under challenging lighting conditions. 
We also plan to exploit the potential of graph neural networks and transformers for other event-based vision algorithms. Another possible extension of the current work could be by integrating the denoising module together with vision algorithms and employing them for robot navigation purposes, autonomous driving cars \cite{1_chen_review}, and healthcare applications such as human fall detection \cite{2_chen_loc}. Eliminating noise events from the observed scene in such scenarios is foreseen to improve the accuracy of the vision algorithms responsible for localizing obstacles and detecting human fall accidents. Noise events, if not eliminated, may be mistaken for real changes in the scene intensities which could results in false positive detections. In the case of autonomous driving, falsely detecting an obstacle along the way will interrupt the vehicle's trajectory and may cause it to take longer paths and more time, which is undesirable. As for human fall detection, noise events may decrease the accuracy of localizing a human and estimating the temporal window for the accident by inflicting erroneous information into the observation. To that end, integrating the proposed denoising method into such systems is envisioned to enhance their accuracy and effectiveness. }
    \vspace{-3.5mm}

\section*{Acknowledgements}

This publication is based upon research work supported by the Khalifa University of Science and Technology under Award No. RC1-2018-KUCARS and the Aerospace Research and Innovation Center (ARIC), which is jointly funded by STRATA Manufacturing PJSC (a Mubadala company) and Khalifa University of Science and Technology. The fourth author of this work, Sajid Javed, is supported by Khalifa University of Science and Technology under the Faculty Start Up grants FSU-2022-003 Award No. 8474000401.

\bibliographystyle{IEEEtran}
\bibliography{IEEEabrv,refD.bib}

\vskip -4\baselineskip

\begin{IEEEbiography}[{\includegraphics[width=1in,height=1.25in,clip,keepaspectratio]{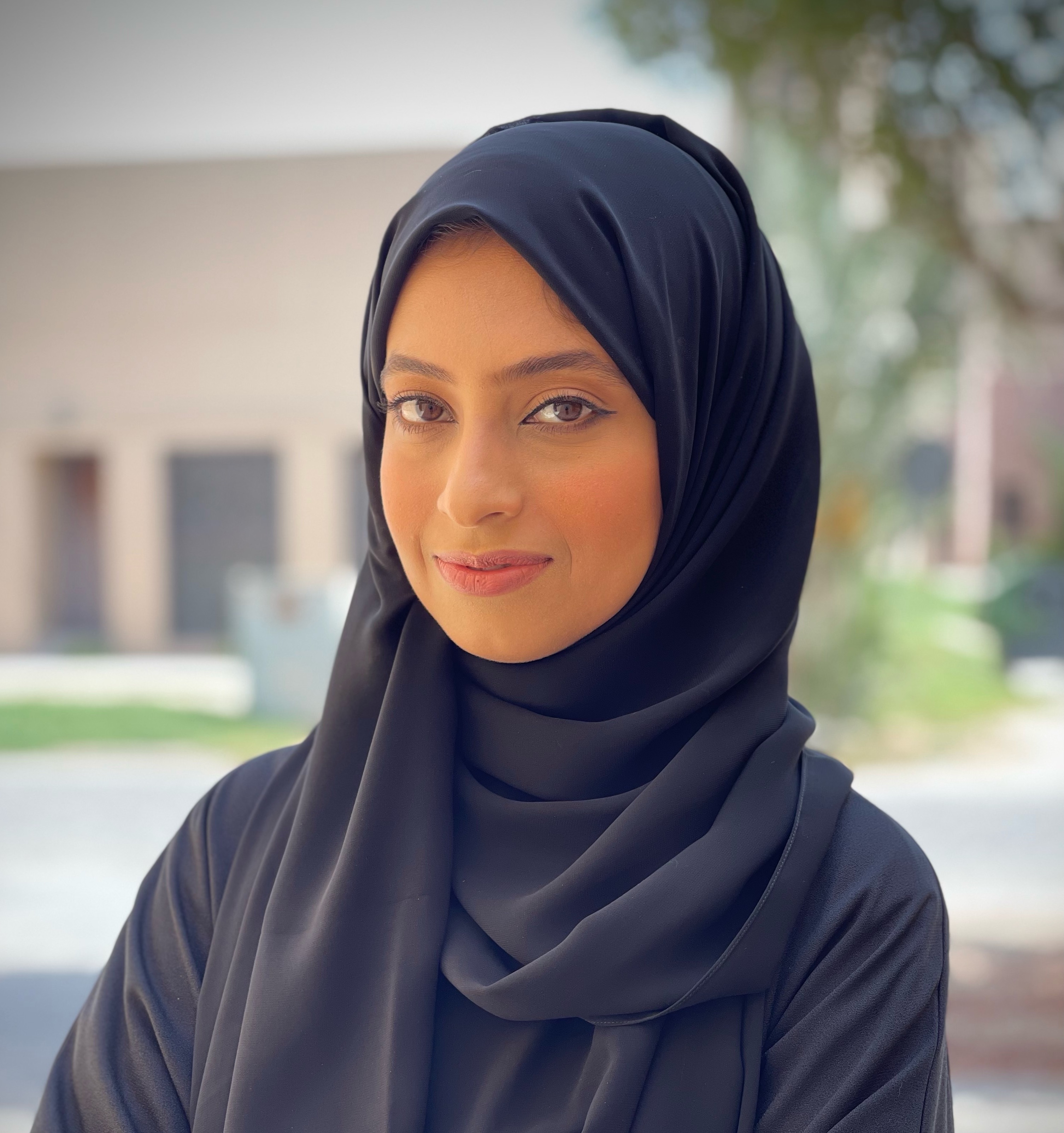}}]{\textbf{Yusra Alkendi}} received the M.Sc. degree in mechanical engineering from Khalifa University, Abu Dhabi, United Arab Emirates, in 2019, where she is currently pursuing the Ph.D. degree in aerospace engineering with a focus on robotics with the Khalifa University Center for Autonomous Robotics Systems (KUCARS). Her current research is focused on the application of artificial intelligence (AI) in the fields of dynamic vision for perception and navigation.\end{IEEEbiography}
\vskip -3\baselineskip 
\vskip -0.5\baselineskip plus -1fil

\begin{IEEEbiography}[{\includegraphics[width=1in,height=1.25in,clip,keepaspectratio]{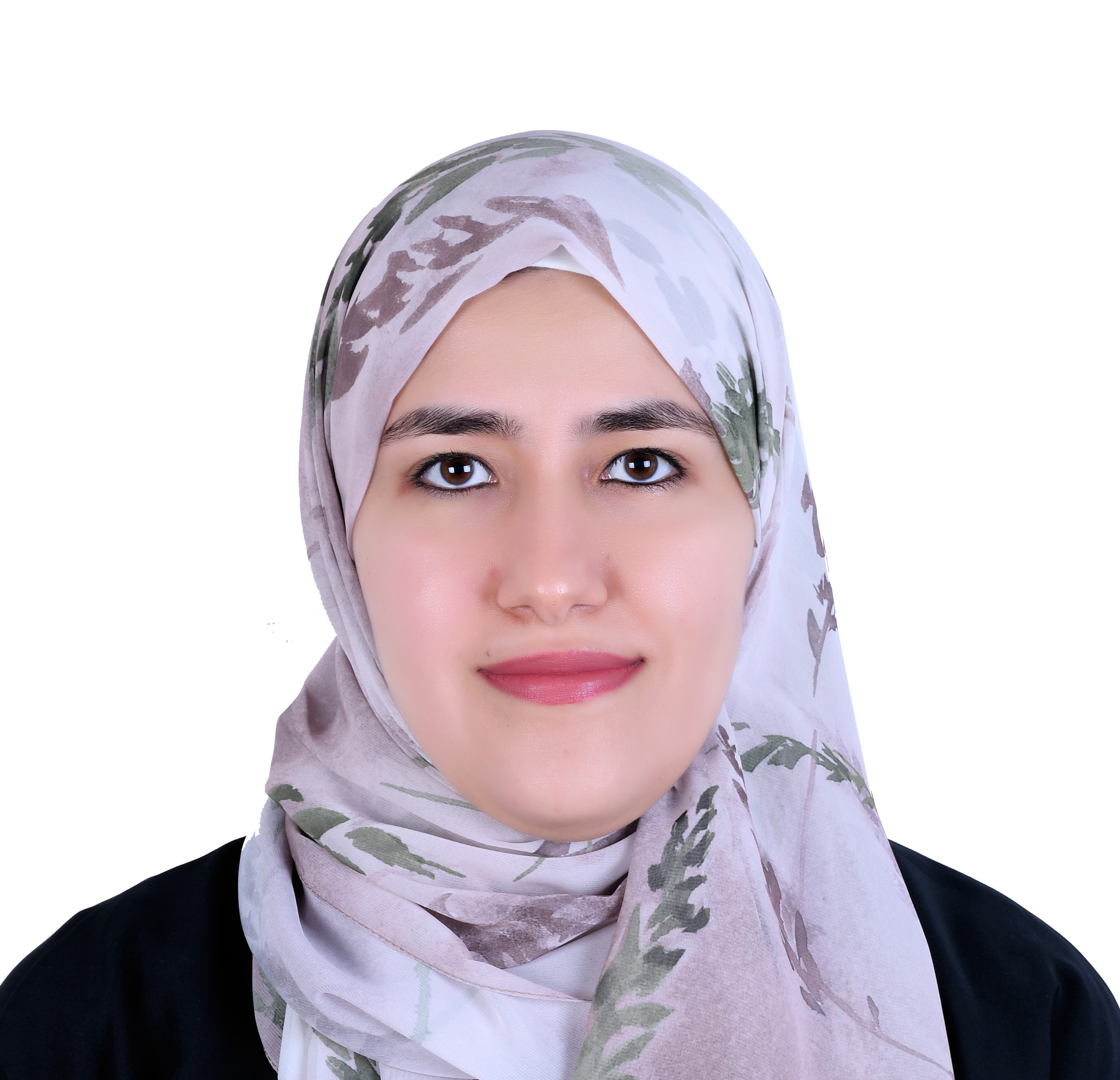}}]{\textbf{Rana Azzam}} received the B.Sc. degree in computer engineering and the M.Sc. degree by Research in electrical and computer engineering from Khalifa University in 2014 and 2016, respectively, and the Ph.D. degree in engineering with a focus on robotics in 2020. She is currently a Postdoctoral Fellow with the Department of Aerospace Engineering. Her research interests include machine learning, reinforcement learning, navigation, and simultaneous localization and mapping.
\end{IEEEbiography}
\vskip -3.5\baselineskip plus -1fil

\begin{IEEEbiography}[{\includegraphics[width=1in,height=1.25in,clip,keepaspectratio]{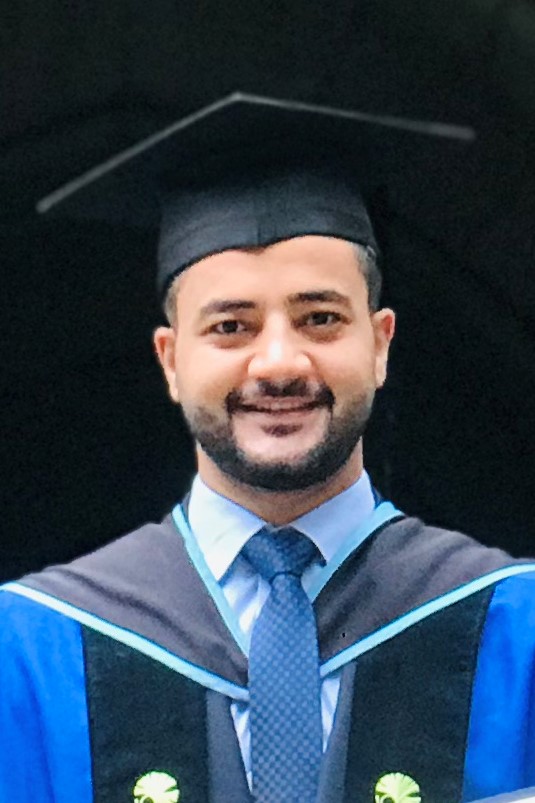}}]{\textbf{Abdulla Ayyad}} (Member, IEEE) received the M.Sc. degree in electrical engineering from The University of Tokyo, in 2019, where he conducted research with the Spacecraft Control and Robotics Laboratory. He is currently a Research Associate with the Khalifa University Center for Autonomous Robotic Systems (KUCARS) and the Aerospace Research and Innovation Center (ARIC) working on several robot autonomy projects. His current research interest includes the application of AI in the fields of perception, navigation, and control.
\end{IEEEbiography}
\vskip -2.8\baselineskip plus -1fil

\begin{IEEEbiography}[{\includegraphics[width=1in,height=1.25in,clip,keepaspectratio]{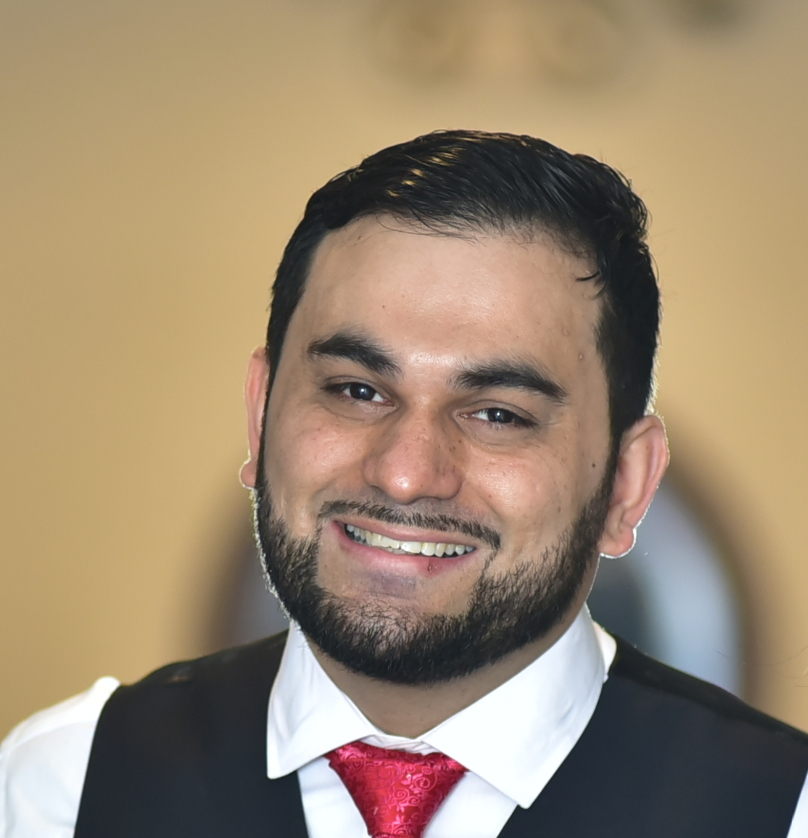}}]{\textbf{Sajid Javed}} is an Assistant Professor of Computer Vision in Electrical Engineering and Computer Science (EECS) department at Khalifa University of Science and Technology, UAE. Prior to that, he was a research scientist at Khalifa University Center for Autonomous Robotics System (KUCARS), UAE, from 2019 to 2021. Before joining Khalifa University, he was a research fellow in the University of Warwick, U.K, from 2017 to 2018, where he worked on histopathological landscapes for better cancer grading and prognostication. He received his B.Sc. degree in computer science from University of Hertfordshire, U.K, in 2010. He
completed his combined Master’s and Ph.D. degree in computer science from Kyungpook National University, Republic of Korea, in 2017. He is also an area chair of ACCV-2022. His research interests include visual object tracking in the wild, multi-object tracking, background-foreground modeling from video sequences, moving object detection from complex scenes, cancer image analytics including tissue phenotyping, nucleus detection, and nucleus classification problems. His research themes involve developing deep neural networks, subspace learning models, grapph neural networks, and vision Transformers.
\end{IEEEbiography}
\vskip -3.2\baselineskip plus -1fil

\begin{IEEEbiography}[{\includegraphics[width=1in,height=1.25in,clip,keepaspectratio]{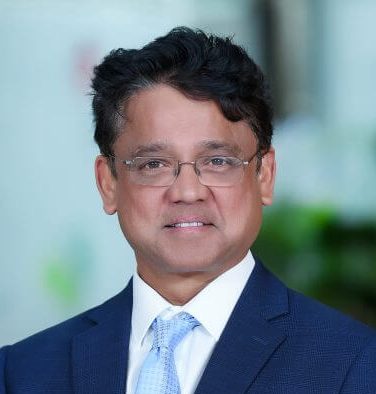}}]{\textbf{Lakmal Seneviratne}} received B.Sc.(Eng.) and Ph.D. degrees in Mechanical Engineering from King's College London (KCL), London,U.K. He is currently a Professor in Mechanical Engineering and the Director of the Robotic Institute at Khalifa University. He is also an Emeritus Professor at King's College London. His research interests are focused on robotics and autonomous systems. He has published over 300 refereed research papers related to these topics. 
\end{IEEEbiography}
\vskip -3.5\baselineskip plus -1fil
\begin{IEEEbiography}[{\includegraphics[width=1in,height=1.25in,clip,keepaspectratio]{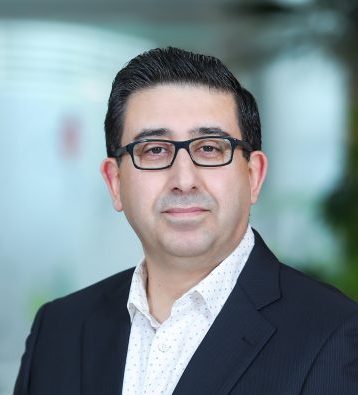}}]{\textbf{Yahya Zweiri}} (Member, IEEE) received the Ph.D. degree from the King’s College London, in 2003. He is currently an Associate Professor with the Department of Aerospace, Khalifa University, United Arab Emirates. He was involved in defense and security research projects in the last 20 years at the Defence Science and Technology Laboratory, King’s College London, and the King Abdullah II Design and Development Bureau, Jordan. He has published over 100 refereed journals and conference papers and filed ten patents in USA and U.K. in unmanned systems field. His research interests include interaction dynamics between unmanned systems and unknown environments by means of deep learning, machine intelligence, constrained optimization, and advanced control.
\end{IEEEbiography}

\vskip -1.8\baselineskip plus -1fil

\vspace{+2mm}

 \appendices

  \section{Additional Qualitative Event Denoising Results} \label{FirstAppendix}
\begin{figure}[hbt]
\centering
\centering
\begin{adjustbox}{width=0.48\textwidth}
  \centering
\begin{tabular}{cccc}

 \includegraphics[width=\columnwidth]{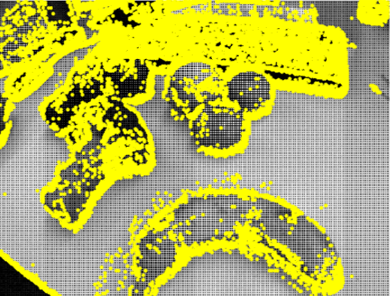} \centering
&         \includegraphics[width=\columnwidth	]{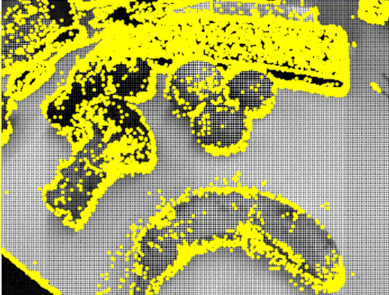} &          \includegraphics[width=\columnwidth]{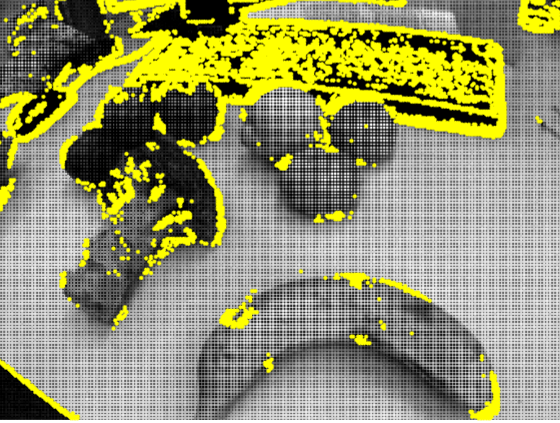} &
\includegraphics[width=\columnwidth]{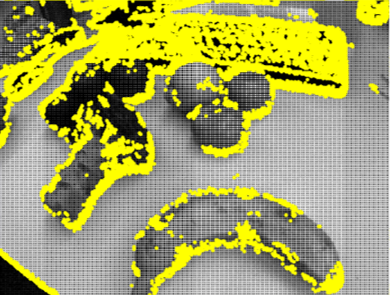}
\\ \multicolumn{4}{c}{\centering\Huge { (a) \textit{SimpleFruit}}}


 \\
 \includegraphics[width=\columnwidth]{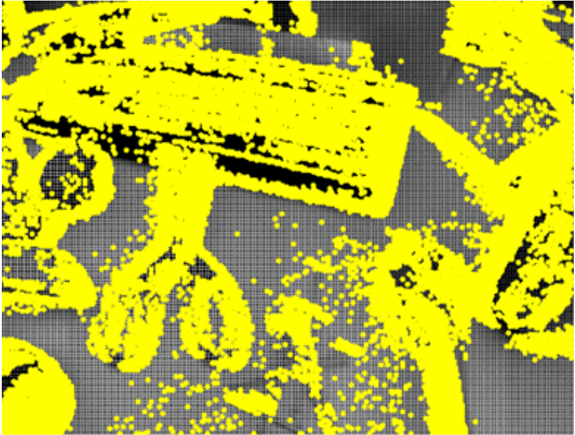} \centering
&         \includegraphics[width=\columnwidth	]{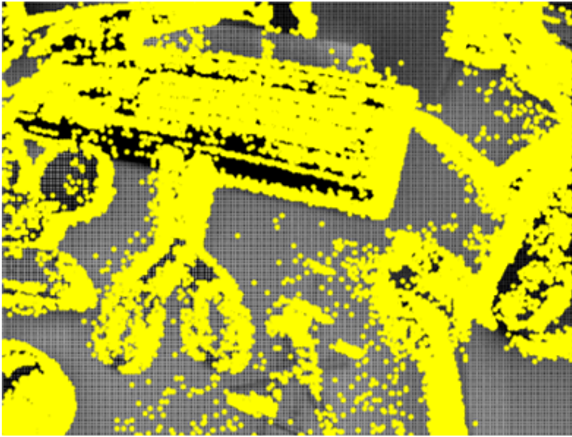} &         \includegraphics[width=\columnwidth]{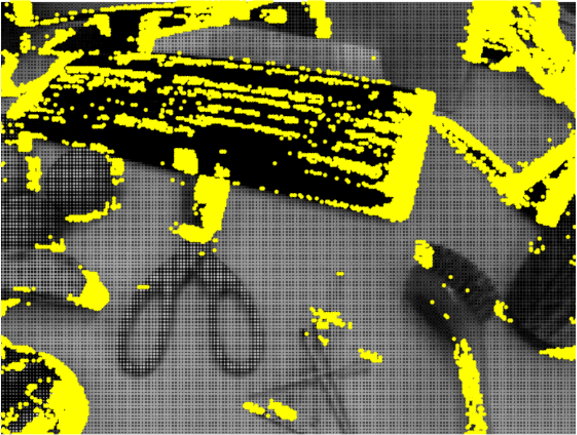} &
\includegraphics[width=\columnwidth]{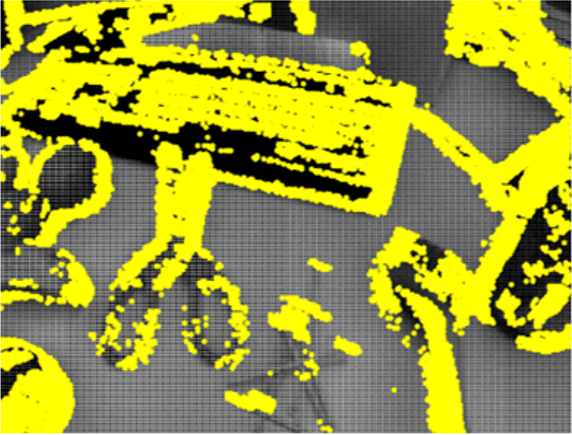} 
\\
\multicolumn{4}{c}{\centering\Huge { (b) \textit{SimpleObjects}}}

\\
\includegraphics[width=\columnwidth]{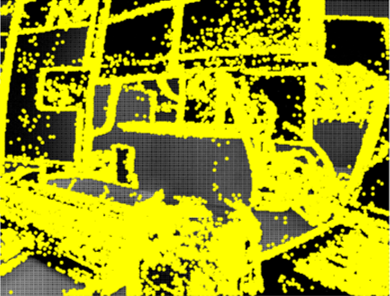}
&         \includegraphics[width=\columnwidth	]{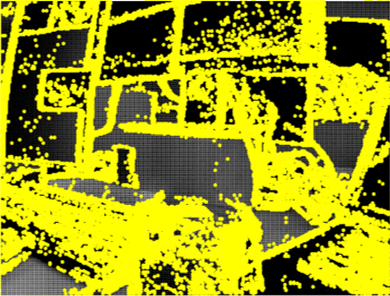} &         \includegraphics[width=\columnwidth]{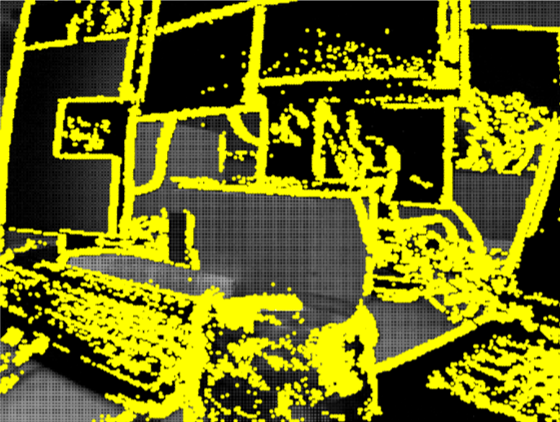} &
\includegraphics[width=\columnwidth]{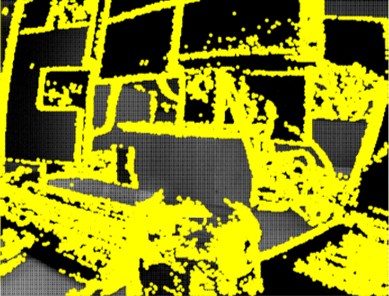}
\\
 \multicolumn{4}{c}{\centering\Huge { (c) \textit{SimpleObjectsDynamic}}}

\\
 \includegraphics[width=\columnwidth]{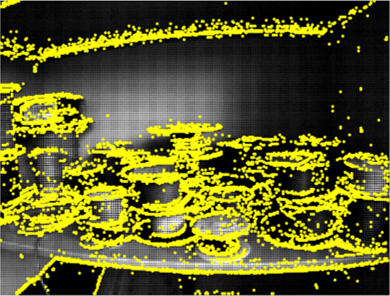} \centering
&         \includegraphics[width=\columnwidth	]{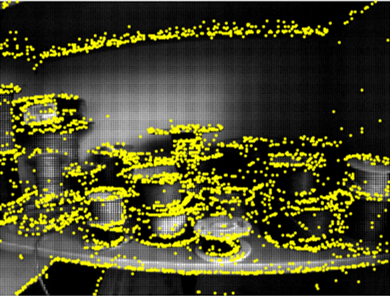} &         \includegraphics[width=\columnwidth]{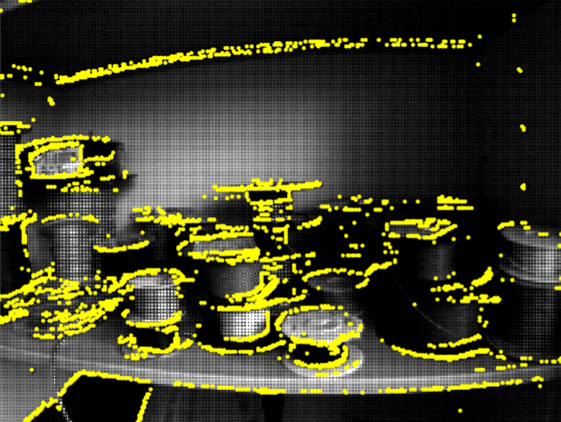} &
\includegraphics[width=\columnwidth]{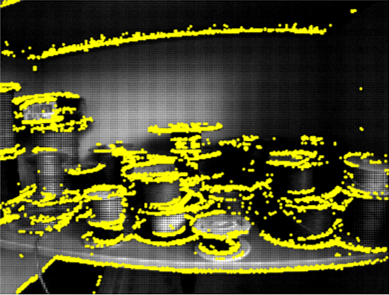}  
\\

\multicolumn{4}{c}{\centering\Huge { (d) \textit{SimpleWires1}}}

\\

 \includegraphics[width=\columnwidth]{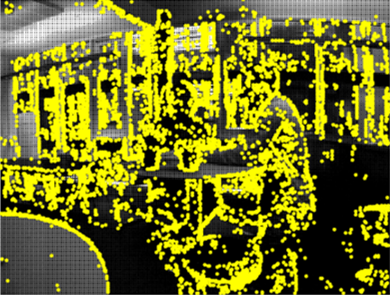} \centering

&         \includegraphics[width=\columnwidth	]{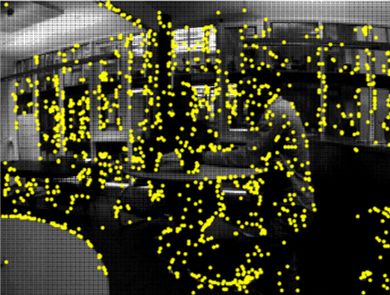} &          \includegraphics[width=\columnwidth]{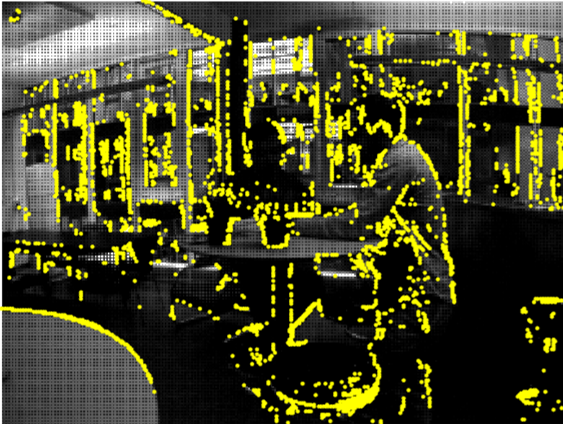} &
\includegraphics[width=\columnwidth]{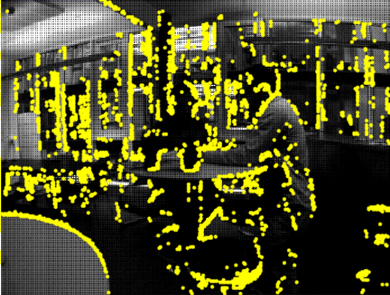}  
\\

\multicolumn{4}{c}{\centering\Huge { (e) \textit{IndoorsKitchen1}}}

\\
 \includegraphics[width=\columnwidth]{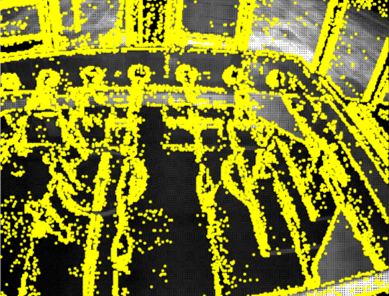} \centering
&         \includegraphics[width=\columnwidth	]{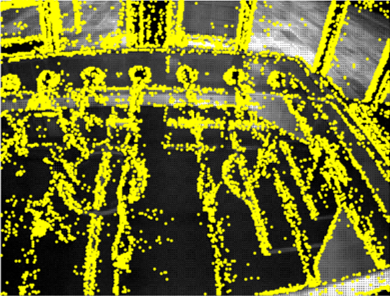} &         \includegraphics[width=\columnwidth]{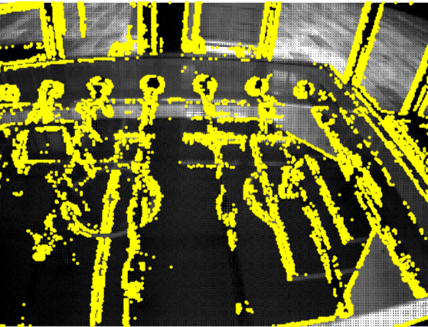} &
\includegraphics[width=\columnwidth]{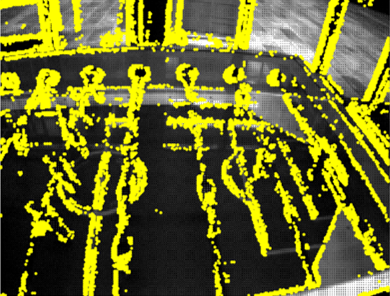} 
\\
\multicolumn{4}{c}{\centering\Huge { (f) \textit{IndoorsFootball1}}}

\\

\includegraphics[width=\columnwidth]{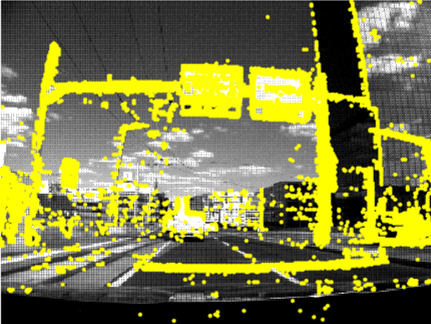} \centering
&         \includegraphics[width=\columnwidth	]{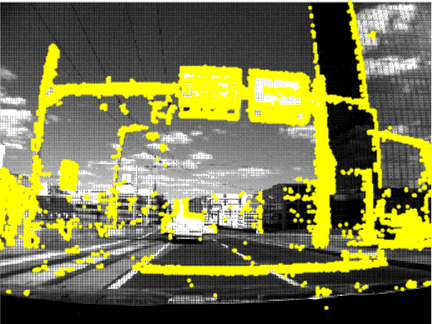} &          \includegraphics[width=\columnwidth]{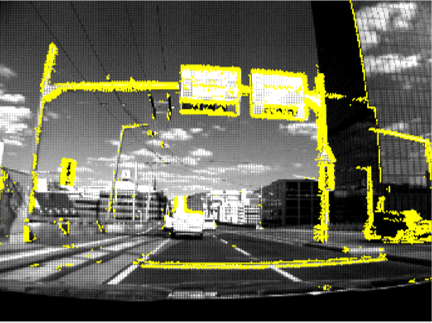} &
\includegraphics[width=\columnwidth]{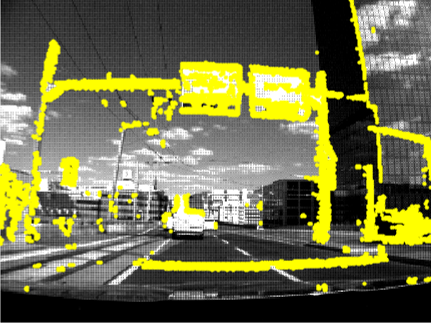} 
\\
\multicolumn{4}{c}{\centering\Huge { (g) \textit{DrivingCity4}}}

\\

 \includegraphics[width=\columnwidth]{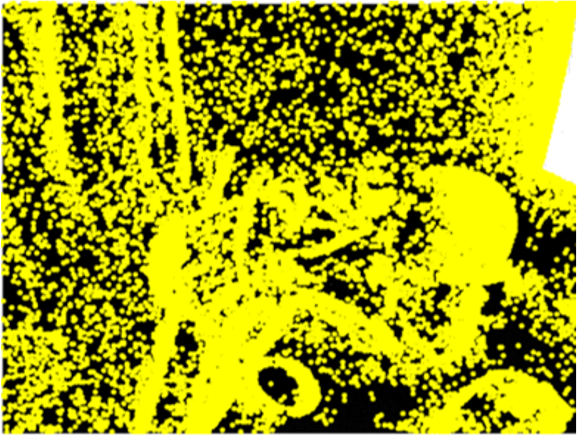} \centering
&         \includegraphics[width=\columnwidth	]{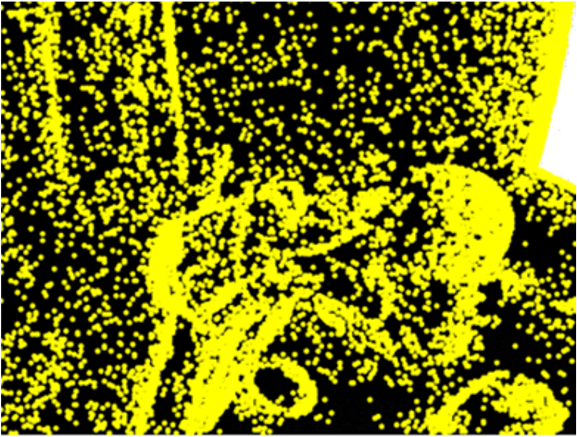} &         \includegraphics[width=\columnwidth]{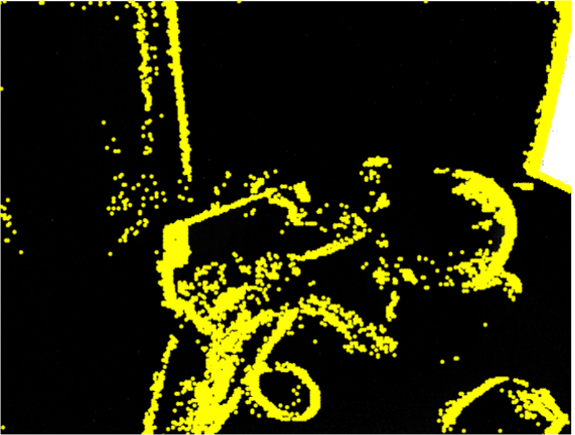} &
\includegraphics[width=\columnwidth]{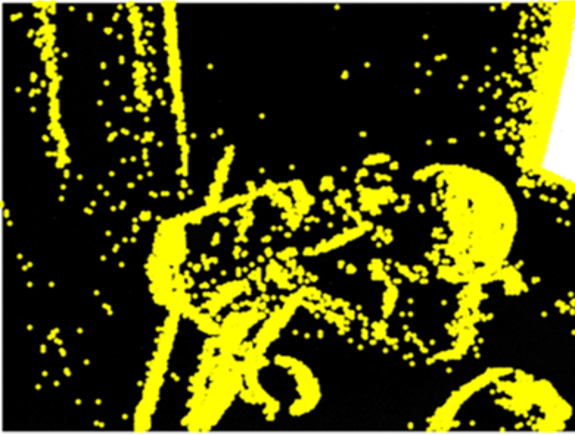} 
\\
\multicolumn{4}{c}{\centering\Huge { (h) \textit{IndoorsDark25ms}}}

\\

          \centering\Huge	 Raw Data - DVS Sensor &  \centering\Huge  Yang filter \cite{D6_Feng2020} & \centering\Huge	{EDnCNN \cite{D9_Baldwin2020a}} 
        &\centering\Huge	{GNN-Transformer (ours)}

\end{tabular}
\end{adjustbox}
\caption{\textcolor{black}{Additional qualitative denoising results tested on published dataset (unseen data), denoised events from DVS (yellow dots) overlaid on APS image.}}\label{qual_3}

\end{figure}

Fig. \ref{qual_3} presents additional qualitative denoising results on other unseen published datasets of our proposed method compared to the state-of-the-art denoising models \cite{D6_Feng2020} and \cite{D9_Baldwin2020a}. 
  
\begin{table}[b] \centering

\caption{\textcolor{black}{Performance comparison of the proposed event denoising classifier and its network variants on the training and testing datasets. 
Note that Case I, Case II, Case III, and Case IV denote GNN, GNN-Transformer 1E1D, GNN-Transformer 2E2D, and GNN-Transformer 3E3D, respectively.
} }
\label{train2test2_2}
\begin{adjustbox}{width=0.49\textwidth,height=0.13\textheight}
\centering
\begin{tabular}{!{\VRule[1.25pt]}c|cccc|c!{\VRule[1.25pt]}}

\specialrule{1.25pt}{0.0pt}{0pt}
\multicolumn{6}{!{\VRule[1.25pt]}c!{\VRule[1.25pt]}}{\centering{Training (Testing) Dataset}}

\\ \hline
\begin{tabular}[c]{@{}c@{}}Event Denoising\\ Model\end{tabular}&
  TP 
&  FP
& TN
 & FN &
\begin{tabular}[c]{@{}c@{}}Filtration \\ $Accuracy$ \end{tabular}
\\\hline
Case I - 3Qs-MSG          & 25750 (5702) & 6250 (2298)  & 23905 (5862) & 8095 (2138)  & 77.59\% (72.28\%) \\
Case I - 4Qs-MSG           & 26008 (5690) & 5992 (2310) & 24046 (6021) & 7954 (1979)  & 78.21\% (73.19\%)\\

Case I - 6Qs-MSG           & 25269 (5315)& 6731 (2685) & 25111 (6213)& 6889 (1787) & 78.72\% (72.05\%)\\

Case I - 7Qs-MSG           & 22446 (5447)& 9554 (2553) & 27329 (6728)& 4671 (1272) & 77.77\% (76.09\%)\\

Case II - 3Qs-MSG & 25745 (5659) & 6255 (2341) & 23998 (5911)& 8002 (2089) & 77.72\% (72.31\%) \\

Case II - 4Qs-MSG & 31353 (7843)& 647 (157)  & 12016 (2939)& 19984 (5061)& 67.76\% (67.39\%)\\

Case II - 6Qs-MSG & 21420 (5394)& 10580 (2606)& 27093 (6702)& 4907 (1298) & 75.80\% (75.60\%)\\

Case II - 7Qs-MSG & 21983 (5497)& 10017 (2503)& 27343 (6742)& 4657 (1258) & 77.07\% (76.49\%)\\

Case III - 3Qs-MSG & 26879 (5731)& 5121 (2269) & 24099 (5938)& 7901 (2062) & 79.65\% (72.93\%)\\

Case III - 4Qs-MSG & 24860 (5167) & 7140 (2833) & 26311 (6556) & 5689 (1444)  & 79.95\% (73.27\%) \\

Case III - 6Qs-MSG & 5168 (1229) & 26832 (6771) & 30972 (7721) & 1028 (279) & 56.47\% (55.94\%)\\

Case III - 7Qs-MSG & 27012 (6403) & 4988 (1597) & 25684 (6513)& 6316 (1487) & \textbf{82.33\%} (\textbf{80.73\%}) \\


Case IV  - 3Qs-MSG & 31428 (7862)& 572 (138)  & 13902 (3469)& 18098 (4531) & 70.83\% (70.82\%) \\
Case IV - 4Qs-MSG & 29157 (7273)& 2843 (727) & 20799 (5161)& 11201 (2839)& 78.06\% (77.71\%)\\
Case IV - 6Qs-MSG & 23377 (5722)& 8623 (2278) & 24081 (6007)& 7919 (1993) & 74.15\% (73.31\%)\\
Case IV - 7Qs-MSG & 18991 (4627)& 13009 (3373) & 28579 (7100)& 3421 (900) & 74.33\% (73.29\%)

\\ \hline  
  \specialrule{1.25pt}{0pt}{0pt}

\end{tabular}

\end{adjustbox}

\label{table5}
\end{table}

\end{document}